\pdfoutput=1

\documentclass[acmsmall]{acmart}

\usepackage{multirow}
\usepackage[mathletters]{ucs}
\usepackage[utf8]{inputenc}
\usepackage[autostyle]{csquotes}
\usepackage{adjustbox}
\usepackage{booktabs}
\usepackage{longtable}
\usepackage{makecell}
\usepackage{caption}
\usepackage{amsmath}
\usepackage{xspace}
\usepackage{comment}

\usepackage{xspace}
\usepackage{comment}
\usepackage{enumitem}

\usepackage{fancyhdr}
\usepackage{soul}

\usepackage{times}
\usepackage{latexsym}
\usepackage{ulem}

\usepackage[T1]{fontenc}

\usepackage[utf8]{inputenc}

\usepackage{microtype}

\usepackage{color} 
\usepackage{listings}

\newcommand{\mquote}[1]{``#1''}
\newcommand{\curlybr}[1]{\{#1\}}

\definecolor{darkspringgreen}{rgb}{0.09, 0.45, 0.27}
\definecolor{jsoncomment}{rgb}{0.5, 0.5, 0.5}
\definecolor{jsonstring}{rgb}{0.0, 0.1, 0.3}
\definecolor{jsonnumber}{rgb}{0, 0, 1}

\lstdefinelanguage{json}{
  basicstyle=\ttfamily,
  commentstyle=\color{jsoncomment},
  stringstyle=\color{jsonstring},
  numberstyle=\color{jsonnumber},
  showstringspaces=false,
  breaklines=true,
  captionpos=b,
  frame=lines,
  literate=
   *{0}{{{\color{jsonnumber}0}}}{1}
    {1}{{{\color{jsonnumber}1}}}{1}
    {2}{{{\color{jsonnumber}2}}}{1}
    {3}{{{\color{jsonnumber}3}}}{1}
    {4}{{{\color{jsonnumber}4}}}{1}
    {5}{{{\color{jsonnumber}5}}}{1}
    {6}{{{\color{jsonnumber}6}}}{1}
    {7}{{{\color{jsonnumber}7}}}{1}
    {8}{{{\color{jsonnumber}8}}}{1}
    {9}{{{\color{jsonnumber}9}}}{1},
}

\begin{document}

\title{ChatGPT as your Personal Data Scientist}

\author{Md Mahadi Hassan}
\email{mzh0167@auburn.edu}
\affiliation{%
  \institution{Auburn University}
  \city{Auburn}
  \state{AL}
}

\author{Alex Knipper}
\email{rak0035@auburn.edu}
\affiliation{%
  \institution{Auburn University}
  \city{Auburn}
  \state{AL}
}

\author{Shubhra Kanti Karmaker Santu}
\email{sks0086@auburn.edu}
\affiliation{%
  \institution{Auburn University}
  \city{Auburn}
  \state{AL}
}

\begin{abstract}

The rise of big data has amplified the need for efficient, user-friendly automated machine learning (AutoML) tools. However, the intricacy of understanding domain-specific data and defining prediction tasks necessitates human intervention making the process time-consuming while preventing full automation. Instead, envision an intelligent agent capable of assisting users in conducting AutoML tasks through intuitive, natural conversations without requiring in-depth knowledge of the underlying machine learning (ML) processes. This agent's key challenge is to accurately comprehend the user's prediction goals and, consequently, formulate precise ML tasks, adjust data sets and model parameters accordingly, and articulate results effectively. In this paper, we take a pioneering step towards this ambitious goal by introducing a ChatGPT-based conversational data-science framework to act as a ``personal data scientist''. Precisely, we utilize Large Language Models (ChatGPT) to build a natural interface between the users and the ML models (Scikit-Learn), which in turn, allows us to approach this ambitious problem with a realistic solution.

Our model pivots around four dialogue states: Data Visualization, Task Formulation, Prediction Engineering, and Result Summary and Recommendation. Each state marks a unique conversation phase, impacting the overall user-system interaction. Multiple LLM instances, serving as ``micro-agents'', ensure a cohesive conversation flow, granting us granular control over the conversation's progression. In summary, we developed an end-to-end system that not only proves the viability of the novel concept of conversational data science but also underscores the potency of LLMs in solving complex tasks. Interestingly, its development spotlighted several critical weaknesses in the current LLMs (ChatGPT) and highlighted substantial opportunities for improvement.

\end{abstract}

\maketitle

\section{Introduction}\label{sec:introduction}

\begin{figure*}
    \centering
    \includegraphics[width=0.85\textwidth]{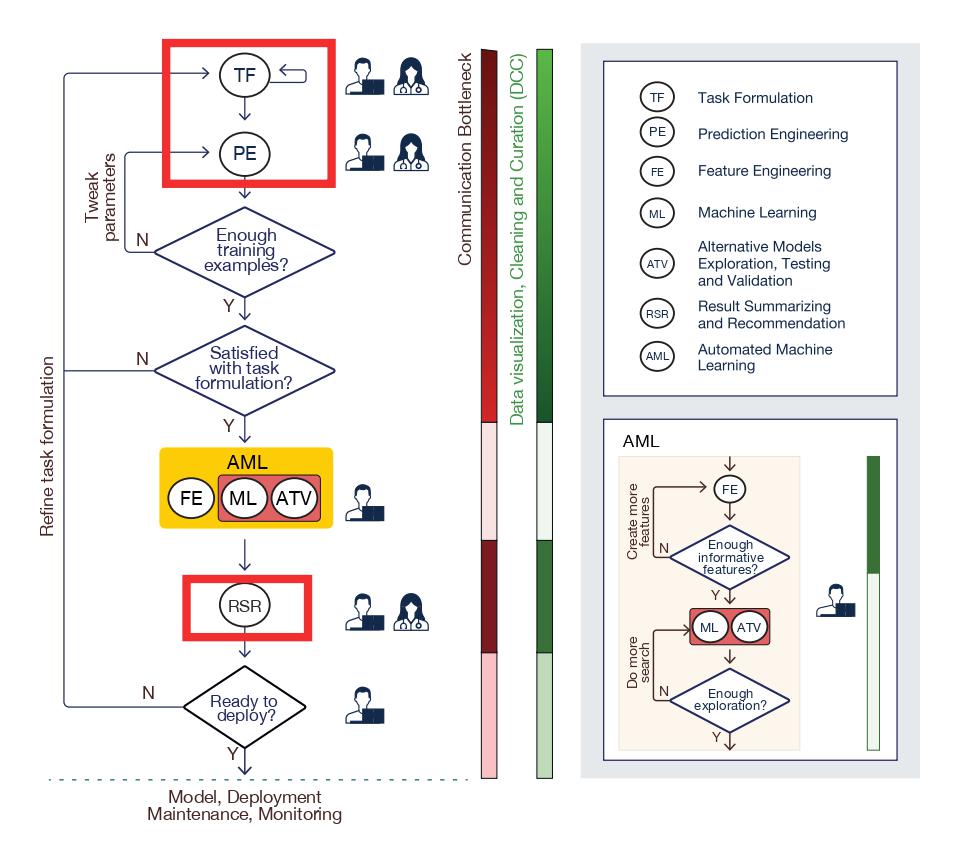}
    \caption{\citet{karmaker2021automl}: A flowchart showing the machine learning process. This chart highlights points of interaction between domain experts and data scientists, along with bottlenecks. In this paper, we focus on automating three steps in the chat cycle with the largest communication bottleneck: Task Formulation (TF), Prediction Engineering (PE), and Result Summarization and Recommendation (RSR).}
    \label{fig:cycle}
\end{figure*}

Automatic Machine Learning (AutoML) tools aim to make machine learning accessible for non-machine learning experts (domain experts), improve the efficiency of machine learning, and accelerate machine learning research. However, the current AutoML process still requires a staggering amount of human involvement at a number of vital steps, as shown in Figure \ref{fig:cycle}. For example, a typical AutoML user would be expected to: 1) Deeply understand the data at their disposal, 2) Know how to create training/testing sets from their data, and 3) Select a promising machine learning technique suitable for their goals. But domain experts (experts in a particular domain other than machine learning working with big data) often lack these understandings and rely on someone well-versed in data science, e.g., a data scientist, to do these tasks~\cite{sarkar2023ad}. These things often still require a prolonged back-and-forth between the domain expert (end-user) and the data scientist. This makes the process rather inefficient for both parties involved and keeps so-called ``AutoML systems'' from being truly automatic~\cite{karmaker2021automl}.

The overall goal of this work is to streamline this lengthy back-and-forth process by making use of a conversational agent which will facilitate the democratization of data science across a wider range of audiences. By doing this, the AI system will be able to help guide users  to express their analytics goals via natural conversation and, subsequently, translate the goal into a well-defined machine learning problem, which, once done, can be automatically executed and interpreted by existing AutoML solutions. Our proposed system, which we will henceforth refer to as VIDS - a \mquote{Virtual Interactive Data Scientist} - aims to be the first of its kind: a true AutoML pipeline and a generalized, well-versed data science assistant. This innovative solution will help lead to the establishment of a higher level of autonomy for AutoML systems, where end-to-end automation is achieved by interfacing large language models with existing AutoML solutions. To be more specific, our dialog-based system aspires to reach the apex of automation, echoing the concept of a Level 6 AutoML system as outlined by \citet{karmaker2021automl}.  This high level of automation, enabled by a consistent, intuitive dialogue with the user, oversees the end-to-end machine learning process, from initial task formulation to comprehensive interpretation of results and subsequent recommendations.

There's no denying the complexity of the task at hand—automating a technical and loosely structured dialogue while concurrently extracting essential information to formulate a cogent machine learning problem. Some critics might view this endeavor as overly ambitious or even unrealistic. However, with the advent of various large language models (LLMs)\cite{palm,gpt3,LLaMA,GLM-130B,opt,RLHF}, such as ChatGPT~\footnote{https://openai.com/blog/chatgpt}, this problem becomes demonstrably more feasible.
These larger models have become very proficient in providing personalized guidance tailored to each user's specific context, ensuring that individual concerns are addressed and any unknown outcomes are effectively explained and interpreted - a level of personalized support that is challenging to achieve with traditional tools. As ChatGPT and similar LLMs continue to evolve, we foresee a future where these models are closely integrated with various industries \& applications, helping to automate tasks, enhance decision-making processes, and assist users in exploring new avenues of innovation.

In this context, the potential of LLMs like ChatGPT extends to even more complex use cases, allowing users to intuitively express their needs and engage in meaningful conversations with an automated system. This affords the potential for creating seamless natural language interfaces for various complex systems, like the aforementioned conversational agent.  If done well, this potential greatly simplifies the automation of interactions between the user and the system. By harnessing the potential of these LLMs, we aim to realize the full VIDS system, revolutionizing the way users interact with and benefit from data science \& machine learning, thereby making these technologies available to a far broader audience. Potential use cases of VIDS become more compelling in dynamic situations where hands-free tools are essential, such as driving, cooking, or battlefield scenarios. This ``natural conversation'' solution allows users to interact with automated machine learning pipelines safely and effectively in these dynamic situations, as well as provides a more ``human'' way to interact with data at scale. Furthermore, the conversational aspect helps accommodate users who may not have complete knowledge of the underlying data and/or have limited access to it. The dialogue naturally helps users understand what tasks are feasible for what reasons and helps them make informed decisions when working with their data.

In summary, breakthroughs in NLP research and language understanding through LLMs, such as ChatGPT, have equipped us with viable technology to realize our ambitious goal of automating the machine learning pipeline through conversational data science. Our solution (VIDS) offers a communication interface that supports natural conversations, enhanced accessibility, personalized guidance, adaptability to dynamic situations, and accommodation for users with limited data knowledge. This innovative solution will empower individuals from diverse backgrounds to harness the power of advanced machine-learning pipelines with ease. As we move forward, these advancements open up the possibility of introducing a new paradigm that will utilize LLMs like ChatGPT to build a virtual interactive data scientist, revolutionizing the way users interact with and benefit from data science and machine learning.

\section{Related Works}

\subsection{Large Language Models}

Large Language Models (LLMs)~\cite{palm,gpt3,LLaMA,GLM-130B,opt,RLHF} have been increasingly recognized as powerful tools in dialog systems. They are widely applied due to their ability to generate human-like text, understand complex language patterns, and provide contextually appropriate responses.

In the context of dialog systems, GPT-3, developed by OpenAI, has been a prominent example in recent literature \cite{gpt3}. It demonstrated significant improvements over its predecessors in terms of fluency and context understanding. By leveraging a Transformer-based architecture, it's able to generate more coherent and contextually appropriate responses compared to earlier models \cite{task-oriented-ds,DLGNet-task}.

Another relevant research area is the application of LLMs for multi-turn dialogues. Here, models like DialoGPT have shown promising results in maintaining conversational context over extended interactions \cite{DialoGPT-sigdial}. They operate by refining the previous response generation models to better maintain the context of the conversation, which significantly enhances the coherence and relevancy of their responses.

Fine-tuning of LLMs for specific domains or tasks within dialog systems is another active area of research. Several studies have focused on techniques such as prompt engineering, rule-based post-processing, or incorporating external knowledge into these models to increase their efficiency and accuracy \cite{Dialog-model-finetuning,FrugalGPT-red-cost}.

Recent works have also begun exploring the integration of LLMs into larger dialog system architectures. For example, studies on systems like HuggingGPT have examined how these models can be leveraged to act as a controller for other AI models \cite{HuggingGPT}.

However, despite the progress made, challenges remain in managing the complexity of multi-turn conversations, ensuring the consistency of responses, and mitigating the tendency of LLMs to generate implausible or \mquote{hallucinated} information \cite{Eval-of-Multi-Turn-DS,vlasov2020dialogue,ReCosa-context-detect,beyond-goldfish-memory}. Therefore, further research is needed to optimize the use of LLMs in dialog systems.

\subsection{Dialog Systems} 

In Dialog Systems research, significant progress has been achieved through advancements in Conversation Topic Prediction~\cite{dialog-topic-modeling} and Dialogue State Tracking (DST)~\cite{henderson2014second,henderson2014third}. DST improvements involve a range of approaches, including schema guidance for better structure~\cite{DBLP:conf/aaai/0002LWZT020, BLP:conf/emnlp/ZhuL0020,kapelonis_multi-task_2022}, recursive inference for deeper understanding~\cite{dst-recursive-inference}, generalization and value normalization for more adaptability~\cite{DBLP:conf/sigdial/Williams13,DBLP:conf/emnlp/WangGZ20}, zero-shot transfer learning for data efficiency~\cite{DBLP:conf/acl/CampagnaFML20,rabinovich_gaining_2022,sarkar2022exploring}, and attention modulation for improved focus during inference~\cite{veron_attention_2022}.
Open-domain dialogue systems have also seen significant advancements. GODEL's~\cite{godel} grounded pre-training adapts to diverse downstream tasks, FusedChat~\cite{young_fusing_2022} combines task-oriented and open-domain dialogue for natural conversations, \& ChatGPT further enhances conversational agent performance across various applications.

\subsection{AutoML Research} 

The ML community as well as the systems community have put a lot of effort in the past decade into automating different \textit{Data Science} pipelines. Major efforts towards automation include \textit{Data Cleaning and visualization}, \textit{Feature Engineering}, \textit{Learning and Parameter Tuning}, \textit{Alternative Models Exploration, Testing and Validation}.
{
\begin{itemize}[leftmargin=*,topsep=0pt,itemsep=-1ex,partopsep=1ex,parsep=1ex]

\item {\bf Data Cleaning and visualization:} This step involves identifying relevant data, handling missing values,  ``\textit{joining}'' multiple data sets, and creating visualizations for improving the quality of the data set. The \textit{Data Mining} and \textit{Databases} community has spent significant effort to automate this step, which has been nicely summarized in~\cite{ilyas2015trends} and~\cite{chu2016data}.

\item {\bf Feature Engineering:} a Data Scientist would attempt to construct useful (informative) features from raw data. Later, these features can be directly fed to ML models to train them and make predictions. In the past 5 years, a number of efforts have focused on automating ``\textit{Feature engineering}'' (\cite{katz2016explorekit,kanter2015deep,mountantonakis2017linked,van2017automatic,khurana2017feature,kaul2017autolearn,wang2017study}).

\item {\bf Learning and Parameter Tuning:}  These include basic machine learning techniques like decision trees, support vector machines, linear regression, neural networks, etc. which have current implementations like scikit-learn~\cite{scikit-learn}, weka~\cite{witten2002data} etc. Machine learning models often contain multiple hyperparameters whose values are critical to obtaining good performance. Automation efforts for hyperparameter tuning include \cite{bergstra2012random,snoek2012practical,hutter2011sequential,bergstra2011algorithms,bengio2012practical,bergstra2013making,swersky2013multi,maclaurin2015gradient}.

\item {\bf Alternative Models Exploration, Testing, and Validation: } 
Automating the process of selecting models, validating them, and finalizing them is critical to the large-scale deployment of ML models. Major automation efforts in this direction include~\cite{thornton2013auto,feurer2015efficient,swearingen2017atm,zoph2016neural,zoph2017learning,liu2017hierarchical,liu2017progressive,real2018regularized,pham2018efficient,baker2017accelerating,xu2019mlfriend,xu2019mlfriend}.

\end{itemize}
}

\smallskip
\noindent However, it is evident that both communities have been reluctant to automate two of the most crucial tasks: \textit{Task Formulation} and \textit{Prediction Engineering}. One particular reason for such reluctance may be attributed to the human-centric nature of these problems. Indeed, both tasks demand significant human interaction during the process and an interactive dialog with the user is necessary to automate this process.

\section{Model Architecture}


This section delves into pioneering methodology of VIDS, illuminating the intricate interplay between overarching structures and localized nuances. Central to our model are four distinct dialogue states - Data Visualization, Task Formulation, Prediction Engineering, and Result Summary and Recommendation, with each representing a unique phase in the conversation and contributing significantly to the overall user-system interaction. VIDS employs stateless global micro-agents, functioning independently of any state-related data or history, to create an overarching structure that enables fluid transitions throughout the dialogue, irrespective of the specific state. This stateless design ensures a smooth narrative flow and avoids complications of state-dependent biases or entanglements, thus bolstering the versatility and adaptability of our dialogue system. Alongside these global agents, local micro-agents, each tailored to a specific dialogue state, proficiently handle the nuances of user utterances and conversation contexts, facilitating smooth transitions between states in line with the evolving dialogue. VIDS' strength lies in this symbiotic relationship between the global and local micro-agents across the different dialogue states. Through this state-oriented, multi-layered approach, we aim to provide a dynamic, user-friendly, and efficient conversation experience, facilitating a streamlined process for automating the machine learning pipeline and fostering improved interaction between users and data science tools.

\begin{figure*}
    \centering
    \includegraphics[width=0.75\textwidth]{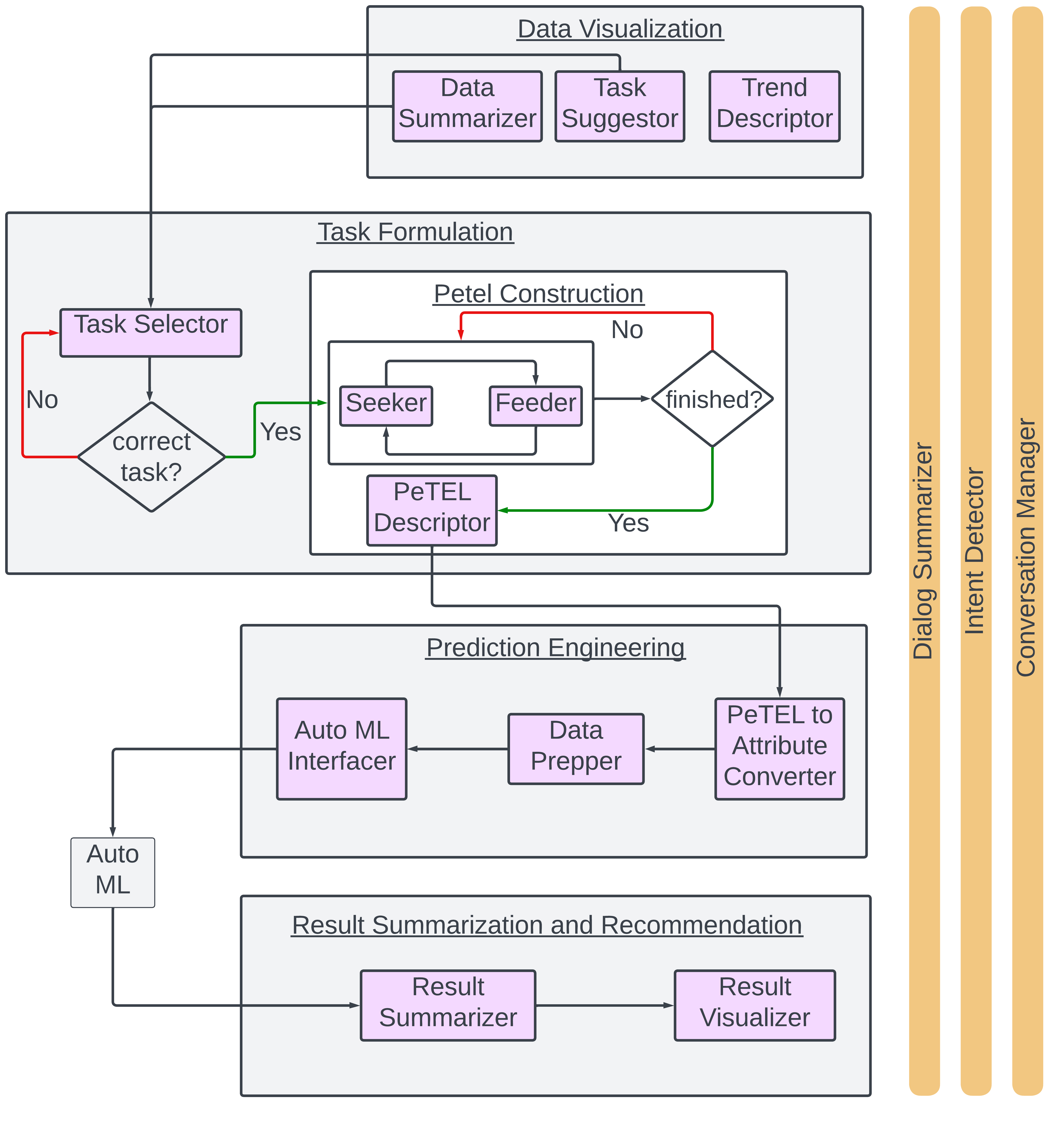}
    \caption{The state diagram of the dialog system. The gray boxes are different state of the conversation history, the dark yellow boxes are global micro-agent, and the purple  }
    \label{fig:stages}
\end{figure*}

\subsection{Global Micro-agents}

\subsubsection{\textbf{State Detector:}}
\label{subsec:micro-state}

The dialog state is a fundamental element, essential in identifying the current phase of the conversation. Its primary task is to ascertain if the user wishes to transition to the next state in the conversation. As shown in Figure~\ref{fig:stages}, VIDS integrates a variety of well-defined states, each corresponding to the different stages of a conversation. The initial state is \mquote{data visualization}, which centers around the presentation of data in a comprehensible and approachable manner. This transitions into the \mquote{task formulation} state, wherein the focus shifts to defining and structuring the task or problem the user wishes to address.
Following this, the system moves into the \mquote{prediction engineering} state. In this phase, the system focuses on constructing and implementing predictive models that are based on the tasks as defined in the previous stage. Finally, the conversation arrives at the \mquote{result summarization and recommendation} state. Here, the system offers a succinct summary of the results, coupled with relevant recommendations based on the outcomes.

The system, considering the immediate context, the current dialog state, and the user's utterance, dynamically determines the user's intent. With this information, the micro-agentor decides whether the user wants to proceed to the next state of the dialog. This approach ensures a smooth flow of conversation, accurately aligning with the user's needs and objectives while offering a user-friendly and engaging experience. The system's design, thus, focuses not only on addressing the user's needs, but also on enriching their interaction with the system. Table~\ref{table:state-prompt} presents the unified prompt design employed to guide ChatGPT to correctly identify the current state of the conversation and the intent of the user.

\newpage
{\small

\begin{longtable}{l|p{12cm}}\hline

    \hline \multicolumn{2}{c}{\textbf{Prompt Design}} \\ \hline 
    \endfirsthead
    
    \multicolumn{2}{c}%
    {{\bfseries \tablename\ \thetable{} -- continued from previous page}} \\
    \hline \multicolumn{2}{c|}{\textbf{Prompt Design}} \\ \hline 
    \endhead
    
    \hline \multicolumn{2}{r}{{Continued on next page}} \\ \hline
    \endfoot

    \endlastfoot

    \multicolumn{2}{l}{\textbf{System setup}} \\* \hline
    \multicolumn{2}{p{14cm}}{ The AI assistant has been designed to understand and categorize user input by detecting the user's intent and conversation state. When processing user input, the assistant must identify the intent from one of the following pre-defined options: `Get dataset info', `Get dataset trend', `Select problem', `Formulate problem', `Problem execution', or `chitchat'. It is essential to note that multiple instances of the same intent type are not permitted. If the user input is unclear or cannot be parsed accurately, the assistant should classify it under the `chitchat' intent, indicating that the input requires further clarification or does not align with the available intent categories. The intent also depends on the current state of the conversation. The rules for state transition are as follows: 
    
    current state: data visualization 
    
    next available states: data visualization, task selection 
    
    current state: ask selection 
    
    next available states: ask selection, task formulation
    
    current state: task formulation 
    
    next available state: task formulation, model training. 
    
    current state: model training 
    
    next available state: model training. 
    
    The agent MUST response with the following json format: \{``intent'': `` '', ``current\_state'': `` '', ``next\_state'': `` '' \}
    } \\* \hline

    \multicolumn{2}{l}{\textbf{Demonstration}} \\* \hline
    User & Taking into account the given context \{ In this dialogue, the AI assistant provided information on suitable machine learning tasks for three different datasets: airlines, flights, and airports. For the flights dataset, the assistant suggested that classification and regression would be suitable tasks. Classification could be used to predict flight delays or cancellations, while regression could be used to predict the amount of delay. The user expressed interest to know more about the dataset.\}, the conversation state \{ dataset\_understanding \} the utterance \{ What details are included in the flight delay dataset? \}, identify my current intent and next state of conversation. Please remember to only response in following format predefined json format without any additional information. Carefully examine the utterance and think about how the context might influence the current utterance, leading you to determine my present intent and next state.\\* \hline
    ChatGPT & \{``intent'': ``Get dataset info'', ``current\_state'': ``dataset\_understanding'', ``next\_state'': ``dataset\_understanding''\}\\* \hline

    User & Taking into account the given context \{ In this dialogue, the AI assistant provided information on suitable machine learning tasks for three different datasets: airlines, flights, and airports. For the flights dataset, the assistant suggested that classification and regression would be suitable tasks. Classification could be used to predict flight delays or cancellations, while regression could be used to predict the amount of delay. The user expressed interest in the flights dataset and asked if it could be formulated as a time series problem, but the assistant did not provide a response to this question. \}, the conversation state \{ dataset\_understanding \} the utterance \{ I want to predict if a flight will be delayed or not \}, identify my current intent and next state of conversation. Please remember to only response in following format predefined json format without any additional information. Carefully examine the utterance and think about how the context might influence the current utterance, leading you to determine my present intent and next state.\\* \hline
    ChatGPT & \{``intent'': ``Select problem'', ``current\_state'': ``dataset\_understanding'', ``next\_state'': ``problem\_selection''\}\\* \hline

    \multicolumn{2}{l}{\textbf{Directive}} \\* \hline

    \multicolumn{2}{p{14cm}}{ Taking into account the given context \{context\}, the conversation state \{conversation state\} the utterance \{user input\}, identify my current intent and next state of conversation. Please remember to only response in following format predefined json format without any additional information. Carefully examine the utterance and think about how the context might influence the current utterance, leading you to determine my present intent and next state.} \\* \bottomrule

\caption{The details of prompt design for the Intent and State Detector micro-agent. In the prompt, the \{context\}, \{conversation state\}, and \{user input\} are placeholders which will be replaced dynamically in different stage of conversation}
    \label{table:state-prompt}
\end{longtable}

}

\subsubsection{\textbf{Dialogue Summarizer:}}
\label{subsec:micro-dialog-summ}

This micro-agent generates concise summaries of the ongoing conversation, enabling effective communication between different micro-agents. By considering the latest user utterance, previous conversation history, and the current response from a micro-agent, this component creates a new dialogue summary that maintains coherence and context throughout the conversation. Table~\ref{table:dialog-prompt} presents the unified prompt design employed to guide ChatGPT to summarize interactions between the user and VIDS.

{\small
\begin{longtable}{p{13cm}}\hline

    \hline \multicolumn{1}{c}{\textbf{Prompt Design}} \\ \hline 
    \endfirsthead
    
    \multicolumn{1}{c}%
    {{\bfseries \tablename\ \thetable{} -- continued from previous page}} \\
    \hline \multicolumn{1}{c|}{\textbf{Prompt}} \\ \hline 
    \endhead
    
    \hline \multicolumn{1}{r}{{Continued on next page}} \\ \hline
    \endfoot

    \endlastfoot

    \textbf{System setup} \\* \hline
     Given the dialog between user and assistant, the AI assistant summarizes the dialog summary. The AI agent should not leave out any crucial information. The goal of this summary generation is not being precise, rather the goal should be to contain all crucial information. if the previous dialog is empty then you should return the current user utterance. \\* \hline

     \textbf{Directive} \\* \hline

    Summarize the following dialog. You should not exclude any important information. \{history\} \\* \bottomrule

\caption{The details of prompt design for the Dialogue Summarizer microprocess. In the prompt, the \{history\} is a placeholders which will be replaced dynamically during the conversation}
    \label{table:dialog-prompt}
\end{longtable}
}

\subsubsection{\textbf{Conversation Manager:}}
\label{subsec:micro-conv-manager}

The conversation management micro-agent integrates input from the appropriate micro-agents to create a coherent, overarching dialogue. This component ensures a seamless user experience and effective task execution by maintaining the dialogue's structure and context throughout the conversation. Table~\ref{table:flow-prompt} presents the unified prompt design employed to guide ChatGPT.

{\small
\begin{longtable}{p{14cm}}\hline

    \hline \multicolumn{1}{c}{ \textbf{Prompt Design}} \\ \hline 
    \endfirsthead
    
    \multicolumn{1}{c}%
    {{\bfseries \tablename\ \thetable{} -- continued from previous page}} \\
    \hline \multicolumn{1}{c}{\textbf{Prompt Design}} \\ \hline 
    \endhead
    
    \hline \multicolumn{1}{r}{{Continued on next page}} \\ \hline
    \endfoot

    \endlastfoot

     \textbf{System setup} \\* \hline
     The AI assistant serves as a virtual data scientist, designed to engage with users and comprehend their objectives. The purpose of this interaction is to develop a machine learning task tailored to the user's data. To achieve this, the assistant will collaborate with various micro agents, each performing specialized tasks to support the primary agent. The assistant will receive context, utterances, dataset summaries, and micro agent responses as input, and should aim to steer the conversation towards the goal. The following micro agents will aid the assistant, providing their output as input to the AI agent for further processing and integration. Depending on the current conversation state, different micro agents will be activated to provide their respective responses: 
    
    Intent Detector: Identifies the user's intent from a list including `Get dataset info', `Get dataset trend', `Select problem', `Formulate problem', `Problem execution', and `Chitchat'. The detected intent will be used to determine the direction of the conversation. 
    
    State Selector: Determines the conversation state, choosing from ``data\_visualization'', ``task\_selection'', ``task\_formulation'', or ``task\_execution''. The chosen state helps the AI agent to adapt its responses and maintain a coherent discussion flow. 
    
    Task Selector: Selects an appropriate ML task from options such as ``classification'', ``regression'', ``clustering'', ``dimensionality reduction'', ``anomaly detection'', and ``time series''. The selected task guides the AI agent in suggesting relevant solutions to the user. 
    
    Task Formulator: Constructs the ML task by utilizing a slot-value filling process. The formulated problem, complete with specified parameters, is then provided to the AI agent, which can assist the user in refining or executing the task. \\* \hline

     \textbf{Directive} \\* \hline

     Taking into account the given context [context], the conversation state \{state\} the utterance \{input\}, current intent \{intent\} and the response from the \{microprocess\} microprocess \{mp\_resp\}, provide appropriate response to the user to carry the conversation to its goal which is formulating a ML task based on user demands. \\* \bottomrule

\caption{The details of prompt design for the Conversation Manager microprocess. In the prompt, \{state\}, \{input\}, \{microprocess\}, and \{mp\_resp\} are placeholders which will be replaced dynamically during the conversation.}
    \label{table:flow-prompt}
\end{longtable}
}

\subsection{Data Visualization}

The interaction pathway of VIDS commences with the Data Visualization stage. Here, users are presented with the option to upload their dataset or choose from an array of pre-existing demonstration datasets. This flexibility fosters an environment of exploration and discovery, enabling users to engage with datasets that align with their specific interests and requirements.

Once a dataset is selected, VIDS embarks on a two-step process to unlock valuable insights from the data. Initially, the system generates a condensed version of the dataset, a maneuver designed to optimize computational resources and streamline subsequent processing. The next step leverages the power of ChatGPT, guided by finely-tuned prompts, to dive deep into the dataset and extract a wealth of insights.

These insights, extracted via the Dataset Summarizer micro-agent, offer users a comprehensive understanding of the dataset, including its overall structure, individual row and column descriptions, and potential visualization ideas. Simultaneously, the Task Suggestor micro-agent analyzes the dataset summary to propose suitable Machine Learning tasks. These interconnected micro-agents ensure a seamless and informative exploration of the dataset, setting the stage for the next phase of interaction.

\subsubsection{\textbf{Dataset Summarizer micro-agent:}}
\label{subsec:micro-dataset-summ}

The Dataset Summarizer micro-agent functions as the heart of the Data Visualization stage. Utilizing a precisely designed prompt, it delves into the reduced version of the dataset, extracting a range of insights that provide users with a comprehensive understanding of the dataset's content, structure, and potential applications. The unified prompt design, presented in Table~\ref{table:dataset-prompt}, guides ChatGPT in this extraction process to ensure the data analysis is thorough and user-friendly.

{\small
\begin{longtable}{p{14cm}}\hline

    \hline \multicolumn{1}{c}{\textbf{Prompt Design}} \\ \hline 
    \endfirsthead
    
    \multicolumn{1}{c}%
    {{\bfseries \tablename\ \thetable{} -- continued from previous page}} \\
    \hline \multicolumn{1}{c}{\textbf{Prompt Design}} \\ \hline 
    \endhead
    
    \hline \multicolumn{1}{r}{{Continued on next page}} \\ \hline
    \endfoot

    \endlastfoot

     \textbf{System setup} \\* \hline
     You are an AI agent who will provide a conprihensive summary of a given dataset. Your task is to provide a comprehensive summary of a given dataset in a strict \mquote{JSON} format.
    
    The summary MUST include the following informations: 
    
    1. dataset summary: the summary of the given dataset in natural language
    
    2. column: it will list all columns and give a brief description about that column
    
    3. Row: AI agent will select a row at random and describe what the row means in natural language
    
    4. Trend: In natural language the AI agent will write the trends that can be found from the given dataset. 
    
    The response should be in a strict JSON format as follows: \curlybr{``summary'': ``...'', ``columns'': [{``name'': ``col1'', ``description'': ``...''}, {``name'': ``col2'', ``description'': ``...''}], ``row'': ``description of a random row'', ``trend'', ``...''}
    
    Please make sure to provide clear and concise descriptions in natural language to facilitate understanding for non-technical users. \\* \hline

     \textbf{Directive} \\* \hline

     Please provide a comprehensive summary of the given dataset. The response MUST be in JSON format NOTHING ELSE. Use the following dataset: \{dataset\}. \\* \bottomrule

\caption{The details of prompt design for the Dataset Summarizer microprocess. In the prompt, the \{dataset\} is a placeholders which will be replaced a miniature version of the user provided dataset.}
    \label{table:dataset-prompt}
\end{longtable}
}

\subsubsection{\textbf{Task Suggestor micro-agent:}}
\label{subsec:micro-task-sugg}

The Task Suggestor micro-agent complements the Dataset Summarizer by proposing suitable Machine Learning tasks based on the dataset summary. This micro-agent employs a unified prompt design, as illustrated in Table~\ref{table:task-suggest}, to guide ChatGPT in generating effective task suggestions. This task suggestion capability enriches the Data Visualization stage, effectively laying the groundwork for the subsequent Task Formulation stage.

{\small
\begin{longtable}{p{14cm}}\hline

    \hline \multicolumn{1}{c}{\textbf{Prompt Design}} \\ \hline 
    \endfirsthead
    
    \multicolumn{1}{c}%
    {{\bfseries \tablename\ \thetable{} -- continued from previous page}} \\
    \hline \multicolumn{1}{l}{\textbf{Prompt Design}} \\ \hline 
    \endhead
    
    \hline \multicolumn{1}{r}{{Continued on next page}} \\ \hline
    \endfoot

    \endlastfoot

     \textbf{System setup} \\* \hline
     The AI agent must analyze the provided dataset summary and recommend appropriate machine learning (ML) tasks. Based on the summary, column descriptions, row information, and any observed trends, the agent should suggest at least two suitable ML task from the following task list: [``classification'', ``regression'', ``clustering'', ``dimensionality reduction'', ``anomaly detection'', ``time series'']. For each ML task the agent chooses a clear rationale must be provided which may include an explanation of why the chosen task aligns with the dataset, and a concrete example of how the task can be formulated. \\* \hline

     \textbf{Directive} \\* \hline

     Suggest ML tasks based on the following dataset summary: \{summary\}  \\* \bottomrule

\caption{The details of prompt design for the suggest ML task a sub-process of Dataset Summarizer microprocess. In the prompt, the \{summary\} is a placeholders which will be replaced by the dataset summary of the user provided dataset.}
    \label{table:task-suggest}
\end{longtable}
}

\subsection{\textbf{Task Formulation}} 

Following the Data Visualization stage, VIDS proceeds to Task Formulation. This section is broken down into two interconnected components: Task Selection and PeTEL Construction, each managed by specialized micro-agents to ensure a thorough and user-oriented formulation of the machine learning task.

\subsubsection{\textbf{Task Selection:}}

Task Selection is the cornerstone of defining the machine learning task. Drawing from the dataset summary and user objectives, this step generates suitable ML tasks for the user to consider. Users have the freedom to select from the suggested tasks or propose an alternative from a pool of common tasks such as \mquote{classification}, \mquote{regression}, \mquote{clustering}, and more. Throughout the dialogue, the system iteratively refines the user's understanding and requirements until a task is selected.
The Task Selection micro-agent (detailed in Section~\ref{subsec:micro-task-selector}) manages this exchange, guiding the user and ensuring the chosen task aligns with their dataset and objectives. The conversation continues until the user is confident in their task choice, promoting effective problem-solving and better outcomes.

\subsubsection{\textbf{PeTEL Construction:}}

Following task selection, the system employs the Prediction Task Expression Language (PeTEL)~\cite{karmaker2021automl} for concise representation of the selected machine learning task. PeTEL uses slot-value pairs to encapsulate the task's essential components, presenting a precise yet comprehensible task description. A complete PeTEL includes the task's desired outcome and search parameters, offering a clear directive for the subsequent ML task.

The PeTEL Construction micro-agent group (detailed from Section~\ref{subsec:micro-seeker} to Section~\ref{subsec:micro-petel-descriptor}) assists in populating necessary values for PeTEL slots based on the chosen ML task. This iterative process guarantees an accurate representation of user requirements, leading to superior results.

The PeTEL Construction concludes with a comprehensive task representation that is user-specific and efficient for further processing. A sample populated PeTEL, demonstrating the iterative process of filling out the different components, is available in Listing~\ref{lst:petel}.

{\small
\begin{lstlisting}[language=json, caption={Sample populated PeTEL for classification task based on FlightDelay dataset (one of our demo datasets).}, label={lst:petel}]
{
  problem_type: classification,
  target_variable: delay_severity,
  features: [departure_airport, arrival_airport, airline, scheduled_departure_time, scheduled_arrival_time, weather_conditions],
  dataset_size: 10000/Default,
  performance_metrics: [accuracy, precision, recall, f1_score, confusion_matrix],
  validation_method: cross_validation,
  classification_methods: [logistic_regression, decision_tree_classifier, random_forest_classifier, svm_classifier, knn_classifier, xgboost_classifier, naive_bayes],
  data_filters: [
    {column: delay_duration, condition: greater_than, value: 15},
    {column: departure_airport, condition: equals, value: JFK}
  ],
  business_goals: [reduce customer complaints, optimize scheduling, improve airport operations],
  additional_requirements: [robust to outliers, handle class imbalance],
  model_preferences: interpretable
}

\end{lstlisting}

}

\subsubsection{\textbf{Task Selector micro-agent:}}
\label{subsec:micro-task-selector}

The task selection micro-agent guides users through a conversation to identify an appropriate machine learning problem from a pool of available options, while also assisting them in choosing a suitable model tailored to their needs. By understanding their requirements and considering the dataset's characteristics, user's objectives, and the dialog context, the assistant is capable of selecting from an array of model types, such as \mquote{classification}, \mquote{regression}, \mquote{clustering}, \mquote{dimensionality reduction}, \mquote{anomaly detection}, etc. This micro-agent facilitates user engagement and ensures the chosen problem and model align with the dataset's properties and the user's goals, offering personalized recommendations that seamlessly integrate into the micro-agent framework. Table~\ref{table:task-select-prompt} presents the unified prompt design employed to guide ChatGPT to select an appropriate ML task from the conversation summary until the user is fixated on a ML task.

{\small
\begin{longtable}{l|p{12cm}}\hline

    \hline \multicolumn{2}{c}{\textbf{Prompt Design}} \\ \hline 
    \endfirsthead
    
    \multicolumn{2}{c}%
    {{\bfseries \tablename\ \thetable{} -- continued from previous page}} \\
    \hline \multicolumn{2}{c}{\textbf{Prompt Design}}  \\ \hline 
    \endhead
    
    \hline \multicolumn{2}{r}{{Continued on next page}} \\ \hline
    \endfoot

    \endlastfoot

    \multicolumn{2}{l}{\textbf{System setup}} \\* \hline
    \multicolumn{2}{p{14cm}}{ The AI assistant is designed to comprehend the user's needs through conversation and assist them in selecting a suitable machine learning model for formulating a Machine Learning problem. The assistant must choose the appropriate model from the provided list: [``classification'', ``regression'', ``clustering'', ``dimensionality reduction'', ``anomaly detection'']. The assistant should consider the user's problem, requirements, and dataset, which may be found in the dialog context, to recommend the best model tailored to their specific needs.} \\* \hline

    \multicolumn{2}{l}{\textbf{Demonstration}} \\* \hline
    \textbf{User} & I want to predict whether a flight will be delayed or not based on factors like weather conditions and previous delays.\\* \hline

    \textbf{ChatGPT} & \{`model': `classification', `reason': `A classification model can be used to categorize flights as delayed or not delayed based on the input features, such as weather conditions and previous delays.'\} \\* \hline

    \textbf{User} & I need to find groups of flights with similar delay patterns, considering variables like departure airport, airline, and time of day.\\* \hline

    \textbf{ChatGPT} & \{`model': `clustering', `reason': `A clustering model can help identify groups of flights with similar delay patterns by analyzing variables like departure airport, airline, and time of day, without requiring labeled data.'\} \\* \hline

    \multicolumn{2}{l}{\textbf{Directive}} \\* \hline

    \multicolumn{2}{p{14cm}}{ Please choose a most suitable model the conversation context: [context] and my latest utterance \{input\}. The output must be in a strict JSON format: \{``model'': ``model name'', ``reason'': ``your detail reasons for the choice''\}.}  \\* \bottomrule

\caption{The details of prompt design for the Task Selector microprocess. In the prompt, the [context], and \{input\} are placeholders which will be replaced dynamically during the conversation}
    \label{table:task-select-prompt}
\end{longtable}
}

\subsubsection{\textbf{Seeker micro-agent:}}
\label{subsec:micro-seeker}
The Seeker micro-agent, part of the PeTEL Construction micro-agent group, converses with the user to populate the next slot in the PeTEL representation. It effectively guides the user through each unfilled slot, ensuring a complete and accurate task formulation. Table~\ref{table:seeker-prompt} presents the unified prompt design employed to guide ChatGPT for asking questions about a specific unfilled slot from the PeTEL expression effectively.

{\small
\begin{longtable}{l|p{12cm}}\hline

    \hline \multicolumn{2}{c}{\textbf{Prompt Design}} \\ \hline 
    \endfirsthead
    
    \multicolumn{2}{c}%
    {{\bfseries \tablename\ \thetable{} -- continued from previous page}} \\
    \hline \multicolumn{2}{c}{\textbf{Prompt Design}}\\ \hline 
    \endhead
    
    \hline \multicolumn{2}{r}{{Continued on next page}} \\ \hline
    \endfoot

    \endlastfoot

    \multicolumn{2}{l}{\textbf{System setup}} \\* \hline
    \multicolumn{2}{p{14cm}}{ The AI assistant functions as a virtual data scientist, aiming to fill in any missing values in a given JSON object by asking the user about them. In order to not overwhelm the user, the assistant should focus on one unidentified slot at a time. With the JSON object, dataset summary, and the next empty slot provided, the agent should formulate a specific and clear question based on this information to accurately identify the missing value. The dataset summary is as follows: \{dataset\_summary\}} \\* \hline

    \multicolumn{2}{l}{\textbf{Demonstration}} \\* \hline
    \textbf{User} & In order to obtain information about the next unidentified slot in the given JSON object 

    \begin{lstlisting}[language=json]
{
    problem_type: time_series_forecasting,
    target_variable: null,
    forecast_horizon: null, 
    
    business_goals: null, 
    
    granularity: null,
    features: null,
    time_range: null, 

    model_preferences: null, 
    
    performance_metrics: null, 
    
    validation_method: null, 
    
    additional_requirements: null, 
    
    time_series_methods: null, 
    
    data_filters: [ 
        {column: null, condition: null, value: null}, 
        {column: null, condition: null, value: null}
    ]
}

    \end{lstlisting}
    
    The agent should ask the user a clear and specific question. Considering the provided context (User wants to formulate a time series forecasting problem by using the flight delay dataset. He is interacting with the agent and the agent is now asking questions to the user to formulate the problem), please formulate a suitable query to gather the necessary data for the missing slot.\\* \hline

    \textbf{ChatGPT} & based on the provided context and the current state of the JSON object, the next unidentified slot is the \textbf{target\_variables} slot. What specific variable are you trying to forecast using the flight delay dataset? This is also known as the target variable. For example, are you trying to forecast the number of flights that will be delayed, or the total duration of delays in minutes? \\* \hline

    \multicolumn{2}{l}{\textbf{Directive}} \\* \hline

    \multicolumn{2}{p{14cm}}{ In order to obtain information about the next unidentified slot in the given JSON object (\{petel\}), the agent should ask the user a clear and specific question. Considering the provided context (\{context\}), please formulate a suitable query to gather the necessary data for the missing slot.}  \\* \bottomrule

\caption{The details of prompt design for the Seeker microprocess.In the system setup, \{dataset\_summary\} is a placeholder for the current dataset summary user is working with. In the prompt, the \{petel\}, and \{context\} are placeholders which will be replaced dynamically during the conversation}
    \label{table:seeker-prompt}
\end{longtable}
}

\subsubsection{\textbf{Feeder micro-agent:}}
\label{subsec:micro-feeder}
The Feeder micro-agent, another part of the PeTEL Construction micro-agent group, populates the PeTEL representation with user responses. It ensures the user's utterance is correctly interpreted and integrated into the problem formulation. Table~\ref{table:feeder-prompt} presents the unified prompt design employed to guide ChatGPT to fill PeTEL espressions' unfilled slots from the user interactions effectively.

{\small
\begin{longtable}{l|p{12cm}}\hline

    \hline \multicolumn{2}{c}{\textbf{Prompt Design}} \\ \hline 
    \endfirsthead
    
    \multicolumn{2}{c}%
    {{\bfseries \tablename\ \thetable{} -- continued from previous page}} \\
    \hline \multicolumn{2}{c}{\textbf{Prompt Design}} \\ \hline 
    \endhead
    
    \hline \multicolumn{2}{r}{{Continued on next page}} \\ \hline
    \endfoot

    \endlastfoot

    \multicolumn{2}{l}{\textbf{System setup}} \\* \hline
    \multicolumn{2}{p{14cm}}{The AI assistant serves as a virtual data scientist, striving to complete a given JSON object by incorporating missing values through user input. Update the JSON object with the relevant unknown slot, considering that the user may also wish to modify existing filled slots. Ensure that the output is a strictly JSON format. If it is not possible to update the JSON object, return the original input JSON object.} \\* \hline

    \multicolumn{2}{l}{\textbf{Demonstration}} \\* \hline
    \textbf{User} & Using the provided information (I'd like to forecast the total number of flight delays.), please update the appropriate slot in the given JSON object:
    
    \begin{lstlisting}[language=json]
{
    problem_type: time_series_forecasting,
    target_variable: null,
    forecast_horizon: null, 
    
    business_goals: null, 
    
    granularity: null,
    features: null,
    time_range: null, 

    model_preferences: null, 
    
    performance_metrics: null, 
    
    validation_method: null, 
    
    additional_requirements: null, 
    
    time_series_methods: null, 
    
    data_filters: [ 
        {column: null, condition: null, value: null}, 
        {column: null, condition: null, value: null}
    ]
}

    \end{lstlisting}
    
    Ensure that you fill only the slot that is directly related to this information. Your response should consist solely of the updated JSON file. \\* \hline

    \textbf{ChatGPT} & 
    \begin{lstlisting}[language=json]
{
    problem_type: time_series_forecasting,
    target_variables: [flight_delays],
    forecast_horizon: null, 
    
    business_goals: null, 
    
    granularity: null,
    features: null,
    time_range: null, 

    model_preferences: null, 
    
    performance_metrics: null, 
    
    validation_method: null, 
    
    additional_requirements: null, 
    
    time_series_methods: null, 
    
    data_filters: [ 
        {column: null, condition: null, value: null}, 
        {column: null, condition: null, value: null}
    ]
}

    \end{lstlisting} \\* \hline

    \textbf{User} & Using the provided information (for time horizon I would like to use one month I guess.), please update the appropriate slot in the given JSON object:
    
    \begin{lstlisting}[language=json]
{
    problem_type: time_series_forecasting,
    target_variables: [flight_delays],
    forecast_horizon: null, 
    
    business_goals: null, 
    
    granularity: null,
    features: null,
    time_range: null, 

    model_preferences: null, 
    
    performance_metrics: null, 
    
    validation_method: null, 
    
    additional_requirements: null, 
    
    time_series_methods: null, 
    
    data_filters: [ 
        {column: null, condition: null, value: null}, 
        {column: null, condition: null, value: null}
    ]
}

    \end{lstlisting}
    
    Ensure that you fill only the slot that is directly related to this information. Your response should consist solely of the updated JSON file. \\* \hline

    \textbf{ChatGPT} & 
    \begin{lstlisting}[language=json]
{
    problem_type: time_series_forecasting,
    target_variables: [flight_delays],
    forecast_horizon: 1 month, 
    
    business_goals: null, 
    
    granularity: null,
    features: null,
    time_range: null, 

    model_preferences: null, 
    
    performance_metrics: null, 
    
    validation_method: null, 
    
    additional_requirements: null, 
    
    time_series_methods: null, 
    
    data_filters: [ 
        {column: null, condition: null, value: null}, 
        {column: null, condition: null, value: null}
    ]
}

    \end{lstlisting} \\* \hline

    \multicolumn{2}{l}{\textbf{Directive}} \\* \hline

    \multicolumn{2}{p{14cm}}{ Using the provided information (\{input\}), please update the appropriate slot in the given JSON object: \{petel\}. Ensure that you fill only the slot that is directly related to this information. Your response should consist solely of the updated JSON file. } \\* \bottomrule

\caption{The details of prompt design for the Feeder microprocess. In the prompt, the \{input\}, and \{petel\} are placeholders which will be replaced dynamically during the conversation}
    \label{table:feeder-prompt}
\end{longtable}
}

\subsubsection{\textbf{PeTEL Descriptor micro-agent:}}
\label{subsec:micro-petel-descriptor}
The PeTEL Descriptor micro-agent, the final part of the PeTEL Construction micro-agent group, communicates the filled PeTEL expression to the user in natural language. It ensures the user fully understands the formulated task, reinforcing the user-centric approach of VIDS. Table~\ref{table:descriptor-prompt} presents the unified prompt design employed to guide ChatGPT.

{\small
\begin{longtable}{l|p{12cm}}\hline

    \hline \multicolumn{2}{c}{\textbf{Prompt Design}}\\ \hline 
    \endfirsthead
    
    \multicolumn{2}{c}%
    {{\bfseries \tablename\ \thetable{} -- continued from previous page}} \\
    \hline  \multicolumn{2}{c}{\textbf{Prompt Design}} \\ \hline 
    \endhead
    
    \hline \multicolumn{2}{r}{{Continued on next page}} \\ \hline
    \endfoot

    \endlastfoot

    \multicolumn{2}{l}{\textbf{System setup} }\\* \hline
    \multicolumn{2}{p{14cm}}{As the AI agent, your primary objective is to carefully examine the problem illustrated within the json object and then develop a comprehensive, yet clear, natural language description. This explanation should effectively communicate the underlying issue to the reader, ensuring that they can grasp the key aspects and implications of the problem at hand. By providing an elaborative and accessible interpretation, the AI agent will enable a better understanding of the situation for individuals who may not be familiar with the json format or the specific context of the issue.} \\* \hline

    \multicolumn{2}{l}{\textbf{Demonstration}} \\* \hline

    \textbf{User} &  Please provide a clear and concise natural language description of the problem formulation presented here: 

    \begin{lstlisting}[language=json]
{
  problem_type: classification,
  target_variable: delay_severity,
  features: [departure_airport, arrival_airport, airline, scheduled_departure_time, scheduled_arrival_time, weather_conditions],
  dataset_size: 10000/Default,
  performance_metrics: [accuracy, precision, recall, f1_score, confusion_matrix],
  validation_method: cross_validation,
  classification_methods: [logistic_regression, decision_tree_classifier, random_forest_classifier, svm_classifier, knn_classifier, xgboost_classifier, naive_bayes],
  data_filters: [
    {column: delay_duration, condition: greater_than, value: 15},
    {column: departure_airport, condition: equals, value: JFK}
  ],
  business_goals: [reduce customer complaints, optimize scheduling, improve airport operations],
  additional_requirements: [robust to outliers, handle class imbalance],
  model_preferences: interpretable
}

    \end{lstlisting}
    
    \\* \hline

    \textbf{ChatGPT} & This is a time series forecasting problem that aims to predict the total delay duration of flights for the next month. The prediction will be based on various features such as departure and arrival airports, airline, weather conditions, and delay duration. The model should be interpretable and capable of handling seasonality, trends, and nonlinear relationships. The problem also aims to understand the impact of weather on flight delays and to forecast delay duration per airline. The dataset is at hourly granularity and covers the past three years. The performance metrics used to evaluate the model include mean squared error, mean absolute error, and R2 score. The model will be validated using time series cross-validation. The data filters used in this problem include only flights departing from JFK airport and operated by Delta airlines. \\* \hline

    \multicolumn{2}{l}{\textbf{Directive}} \\* \hline

    \multicolumn{2}{p{14cm}}{ Please provide a clear and concise natural language description of the problem formulation presented here: \{petel\}. }\\* \bottomrule

\caption{The details of prompt design for the PeTEL Descriptor microprocess. In the prompt, \{petel\} is a placeholders which will be replaced by a fully filled PeTEL expression.}
    \label{table:descriptor-prompt}
\end{longtable}
}

\subsection{Prediction Engineering}

Following Task Formulation, the journey progresses to Prediction Engineering, a fundamental stage where the system transforms the problem representation into a tangible prediction model. This phase is composed of three primary steps: PeTEL to Feature, Data Cleaning and Preparation, and AutoML interfacing. Each step is crucial in bridging the gap between the problem's conceptual representation and its practical implementation.

\subsubsection{\textbf{PeTEL to Attribute Converter:}}

The PeTEL to Feature conversion is the first step in the Prediction Engineering process. Here, the PeTEL representation, which succinctly describes the machine learning task, is translated into features that can be used by the prediction model. This process ensures that the machine learning algorithms can interpret and work with the problem definition, turning the abstract task representation into concrete, computable features.

\subsubsection{\textbf{Data Prepper Micro-Agent:}}

Once the features are defined, the next step is Data Cleaning and Preparation. This stage involves pre-processing the dataset to ensure it's suitable for the prediction model. Common procedures during this phase include handling missing data, dealing with outliers, and encoding categorical variables. The goal is to produce a clean, well-structured dataset that can be readily consumed by downstream machine learning algorithms, maximizing the potential for accurate and meaningful predictions.

\subsubsection{\textbf{AutoML interfacer Micro-Agent:}}

The final step in the Prediction Engineering phase is interfacing with AutoML systems. AutoML platforms automate the process of applying machine learning to real-world problems, making the technology accessible to non-experts and improving efficiency of experts. In this step, the prepared dataset is fed into an AutoML system, which automatically selects the most suitable machine learning algorithm, optimizes its parameters, and trains the model. The result is a robust prediction model that is ready to generate insights from new data, bringing the conceptual machine learning task to fruition.

\subsection{Result Summary and Recommendation}

A data scientist's work typically culminates in consolidating any findings and suggesting optimal approaches to domain experts. These recommendations can span diverse levels, such as models, features, or computational overhead. However, this crucial stage is primarily manual and lacks systematic structuring in the current landscape.
In response to this, we aim to enhance and refine the final phase of VIDS, known as the Result Summary and Recommendation, in upcoming iterations. We anticipate incorporating two primary processes within this phase: Result Summarization and Result Visualization. These proposed enhancements aim to bolster users' comprehension and capacity to make informed decisions, thereby streamlining the intricate process of data science.

\subsubsection{\textbf{Result Summarizer Micro-Agent:}}

Currently, we have implemented the Result Summarization micro-agent, where the system produces a comprehensive summary of the findings once the machine learning tasks have been executed. Utilizing an AutoML library such as Auto-SKLearn, the system trains all specified models, equipping users with a broad comparison to discern the most effective solution. This process distills the results into an accessible format, enabling users to grasp the essence of the findings quickly.

\subsubsection{\textbf{Result Visualizer Micro-Agent} (Future work):}

Looking forward, we aim to implement the Result Visualization micro-agent. Visualizing the outcomes can significantly aid users' understanding and facilitate more informed decision-making. We plan to develop a process that generates suitable visualizations based on the results, such as performance metrics or feature importance, offering a more intuitive perspective of the findings.

Additionally, we aspire to further optimize the decision-making process, by having the system engage in an interactive dialogue with the user, recommending the most suitable model based on the conversation. This personalized approach would empower users to make informed decisions, streamline the process, and enhance the effectiveness of their machine learning endeavors.

\section{Qualitative Examples}

The purpose of this section is to conduct a thorough investigation of the qualitative aspects of Large Language Models (LLMs) in complex tasks, with a specific focus on three key areas. This study will use the {Student Performance (SP)}\footnote{https://www.kaggle.com/datasets/larsen0966/student-performance-data-set} dataset. First, we will examine the overall chat cycle, which involves the interactions between the user and VIDS, consisting of well-designed assemblies of LLM agents. This analysis will provide insights into the user experience and highlight the usability and effectiveness of the VIDS' LLMs in facilitating seamless communication. Secondly, we will explore the interaction among the micro-agent agents themselves, each equipped with its own LLM instance. This exploration will deepen our understanding of the internal mechanics of the micro-agent agents and their role in executing complex tasks. Lastly, we will conduct a comprehensive evaluation of Prompt Taxonomy testing on these micro-agent agents to assess their adaptability and responsiveness in diverse scenarios, further emphasizing the practicality and usability of LLMs in complex tasks. By conducting this comprehensive qualitative analysis, our goal is to provide a holistic understanding of the role and effectiveness of LLMs in managing complex tasks.

\subsection{Overall Chat Cycle} 

In this subsection, we turn our attention to the overall chat cycle, a crucial component of the interaction between the user and the system. This cycle, facilitated by VIDS' well-designed assembly of Large Language Model (LLM) agents, is the cornerstone of the model's user-system communication. We will delve into the intricacies of this cycle, examining how the LLMs contribute to a seamless and effective communication experience. Our analysis will provide valuable insights into the user experience, highlighting the usability and effectiveness of the system's LLMs in complex tasks.

{  \small
\begin{longtable}{l|p{12cm}}\hline

    \hline {\textbf{Turn}} &  {\textbf{Conversation}} \\ \hline 
    \endfirsthead
    
    \multicolumn{2}{c}%
    {{\bfseries \tablename\ \thetable{} -- continued from previous page}} \\
    \hline \multicolumn{1}{c|}{\textbf{Turn}}  & \multicolumn{1}{c}{\textbf{Conversation}} \\ \hline 
    \endhead
    
    \hline \multicolumn{2}{r}{{Continued on next page}} \\ \hline
    \endfoot

    \endlastfoot

    $system_1$ & Welcome, I am your personal data scientist. I have processed the provided dataset for your convenience I present the summary here: \curlybr{Summary}. I propose the following ML tasks for this dataset: \curlybr{$ML\_task$} \\* \hline
    
    $utterance_1$ & Ok, from the description it seems like classification is a good choice. \\* \hline
    
    $response_1$ & Great! Let's move forward with the classification task. Now that we have decided on the classification task, let's move on to formulating the problem. Can you provide me with more details on what you would like to achieve with this task? For example, what is the target variable you would like to predict? \\* \hline

    $utterance_2$ & I think I will use final grade \\* \hline

    $response_2$ &  Based on the provided context, the next unidentified slot in the JSON object is `dataset\_size'. To obtain this information, the agent could ask the user: Can you please provide the size of your dataset for the Classification model you are building? \\* \hline

    $utterance_3$ & Lets use 10000 samples for this task \\* \hline

    \multicolumn{2}{c}{... ... ...} \\* \hline

    $response_n$ &  The task is to predict the final\_grade of a student based on various features like study\_hours, attendance, participation, homework\_scores, and test\_scores. The dataset size is hypothetical. Performance is measured using accuracy, F1 score, and the confusion matrix. K-fold cross-validation is used for validation. Three classification methods are considered: random forest, SVM, and logistic regression. 
     
     The data is filtered to only include records where attendance is greater than 75 (assuming attendance is measured as a percentage) and study\_hours is more than 1. The business goal is to predict student performance to implement early interventions for those at risk of failing. Model interpretability is highlighted as an important additional requirement, and there's a preference for models with higher accuracy, but interpretability is also considered important. \\* \hline

    $utterance_{n+1}$ & That seems all right to me. go ahead with this task. \\* \hline

\caption{Interaction of user and system where $utterance_*$ are the user utterances and $response_*$ are the overall system response to the user. In $system_1$, \curlybr{summary} and \curlybr{$ML\_task$} are from Table~\ref{table:state-data-vis}.}
    \label{table:overall-cycle-classification}
\end{longtable}

}

\subsection{Interaction Between Micro-Agents}

In this subsection, we delve into the interactions within the micro-agents themselves, each equipped with its own instance of a Large Language Model (LLM). These interactions occur in four distinct states within the system: data visualization, task formulation, prediction engineering, and result generation and summarization. By examining the interplay among these micro-agents in each state, we aim to enhance our understanding of the internal mechanics of the LLMs and their role in executing complex tasks.

\subsubsection{\textbf{Data Visualization State:}}
The first state, data visualization, involves the presentation of data in a visual format that aids in understanding the complexities and patterns within the data. The interaction between micro-agent agents in this state is crucial as it determines the effectiveness of the data representation. In Table~\ref{table:state-data-vis}, we show the specific roles and interactions of the micro-agents in this state.

{\small
\begin{longtable}{l|l|l|p{9cm}}\hline

    \hline {\textbf{\#}} & {\textbf{Input}} & {\textbf{Micro-Agent} } & {\textbf{Output}} \\ \hline 
    \endfirsthead
    
    \multicolumn{4}{c}%
    {{\bfseries \tablename\ \thetable{} -- continued from previous page}} \\
    \hline \multicolumn{1}{c|}{\textbf{\#}} & \multicolumn{1}{c|}{\textbf{Input}} & \multicolumn{1}{c|}{\textbf{Micro-Agent}} & \multicolumn{1}{c}{\textbf{Output}} \\ \hline 
    \endhead
    
    \hline \multicolumn{4}{r}{{Continued on next page}} \\ \hline
    \endfoot

    \endlastfoot

    1 & Dataset & --- & $Min\_ds :$ Miniature dataset \\* \hline
    2 & $Min\_ds$ & Dataset Summarizer & 
                $summary:$ \{
                
                    dataset\_summary : This dataset contains information about students in a school. It includes their demographic information such as sex, age, address, family size, and parents' education level. It also includes information about their academic performance such as grades, study time, and number of failures. Additionally, it includes information about their social life such as alcohol consumption, going out with friends, and having a romantic relationship.

                    columns : list of column names and description

                    sample row : This row represents a female student who is 16 years old, lives in an urban area with a family size greater than 3, and her parents are living together. Her mother has a higher education level than her father and works in the health sector, while her father works in the services sector. She chose this school because of its reputation and her father is her guardian. It takes her less than 15 minutes to travel from home to school. She studies for 2 to 5 hours per week and has not failed any classes in the past. She receives no extra educational support from school or family members, and does not pay for extra classes. She participates in extracurricular activities and attended nursery school. She wants to take higher education and has internet access at home. She is not in a romantic relationship and has a good quality of family relationships. She has a moderate amount of free time and goes out with friends occasionally. She has a low workday alcohol consumption and a very low weekend alcohol consumption. Her current health status is good and she has had 6 school absences. Her grades are 12 in the first period, 12 in the second period, and 13 in the final period.

                    trend : From the dataset, it can be observed that female students tend to have higher grades than male students. Additionally, students who receive extra educational support from school or family members tend to have higher grades than those who do not. Students who have a higher amount of study time per week also tend to have higher grades. On the other hand, students who have a higher amount of alcohol consumption tend to have lower grades. Finally, students who have a higher number of school absences tend to have lower grades.
                \} \\* \hline
    3 & $summary$ & Suggest ML task & $\textbf{ML\_task : }$ Based on the provided dataset summary, the following are the suitable ML tasks:
    
    1. Classification: Classification can be used to predict whether a student will pass or fail based on their demographic information, academic performance, and social life. This task can be formulated as a binary classification problem where the target variable is pass/fail. The features can include sex, age, address, family size, parents' education level, study time, number of failures, alcohol consumption, and school absences.
    
    2. Regression: Regression can be used to predict the final grade of a student based on their demographic information, academic performance, and social life. This task can be formulated as a regression problem where the target variable is the final grade (G3). The features can include sex, age, address, family size, parents' education level, study time, number of failures, alcohol consumption, and school absences.
    
    The rationale for choosing classification is that it can help identify students who are at risk of failing and provide early intervention to improve their academic performance. The rationale for choosing regression is that it can help identify factors that contribute to a student's academic success and provide insights into how to improve their performance. \\* \hline

\caption{Interaction of different micro-agent in the Data Visualization state. In this qualitative study we work with Flight Delay dataset.}
    \label{table:state-data-vis}
\end{longtable}
}

\subsubsection{\textbf{Task Formulation State:}}

The second state, task formulation, is the stage where the task to be performed is defined and structured. The interaction between micro-agent agents in this state is critical as it shapes the formulation of the task and ultimately influences the success of the task's execution. In Table~\ref{table:state-task-formulate}, we delve into the specifics of these interactions.

{\small
\begin{longtable}{l|p{1.5cm}|p{2cm}|p{8cm}}\hline

    \hline {\textbf{\#}} & {\textbf{Input}} & {\textbf{Micro-Agent} } & {\textbf{Output}} \\ \hline 
    \endfirsthead
    
    \multicolumn{4}{c}%
    {{\bfseries \tablename\ \thetable{} -- continued from previous page}} \\
    \hline \multicolumn{1}{c|}{\textbf{\#}} & \multicolumn{1}{c|}{\textbf{Input}} & \multicolumn{1}{c|}{\textbf{Micro-Agent}} & \multicolumn{1}{c}{\textbf{Output}} \\ \hline 
    \endhead
    
    \hline \multicolumn{4}{r}{{Continued on next page}} \\ \hline
    \endfoot

    \endlastfoot

    1 & $utterance_1$, $summary$, $ML\_task$ & Dialog summarizer & $context_1 :$ The dialog provides a summary of a dataset containing information about students in a school, including their demographic information, academic performance, and social life. The dataset summary highlights trends such as female students having higher grades and students who receive extra educational support having higher grades. Based on this information, the user suggests that classification would be a suitable machine learning task to predict whether a student will pass or fail based on their demographic information, academic performance, and social life. The user's rationale for choosing classification is to identify students who are at risk of failing and provide early intervention to improve their academic performance.\\* \hline
    2 & $context_1$, $utterance_1$ & State Selector & $ds_1 :$ Task Formulation \\* \hline
    3 & $utterance_1$, $context_1$, $ds_1$ & Task Selector & $ML\_task : $ 
                \curlybr{ \mquote{model}: \mquote{classification}, \mquote{reason}: \mquote{Based on the dataset summary, the AI agent suggested two suitable machine learning tasks: classification to predict whether a student will pass or fail, and regression to predict a student's final grade. The user agreed that classification would be a good choice based on the dataset summary. Therefore, a classification model is a good choice for this task.}}    \\* \hline
    4 & $utterance_2$, $context_1$, $ds_1$ & Dialog Summarizer & $context_2 :$ In this dialogue, the AI assistant provides a summary of a dataset containing information about students in a school, including their demographic information, academic performance, and social life. The assistant notes that female students tend to have higher grades than male students, and students who receive extra educational support from school or family members tend to have higher grades. The assistant suggests two suitable ML tasks for the dataset: classification and regression. The user decides to use classification to predict whether a student will pass or fail based on their demographic information, academic performance, and social life. The assistant asks for more details on the target variable, and the user decides to use the final grade as the target variable.\\* \hline
    5 & $context_2$, $utterance_2$ & State Selector & $ds_2 :$ Task Formulation \\* \hline

    6 & $utterance_2$, $context_2$, $PeTEL_1$ & Feeder & $PeTEL_2 :$ 
    \begin{lstlisting}[language=json]
{
    problem_type: classification, 
    target_variable: final grade, 
    features: None, 
    dataset_size: None, 
    performance_metrics: None, 
    validation_method: None, 
    classification_methods: None, 
    data_filters: [
        {column: final grade, condition: less than, value: passing grade}, 
        {column: None, condition: None, value: None}
        ], 
    business_goals: None, 
    additional_requirements: None, 
    model_preferences: None

}
    \end{lstlisting}

    \\* \hline

    7 & $utterance_2$, $context_2$, $PeTEL_2$ & Seeker & $Q_1 :$ Based on the provided context, the next unidentified slot in the JSON object is `dataset\_size'. To
obtain this information, the agent could ask the user: Can you please provide the size of your dataset for the Classification model you are building?\\* \hline
    
    8 & $utterance_3$, $context_2$, $Q_1$ & Dialog Summarizer & $context_3 :$ 
    The AI assistant provided a summary of the dataset, including information about the columns and a sample row. The assistant also mentioned trends observed in the dataset, such as the correlation between study time and grades, and the negative correlation between alcohol consumption and grades. The user asked for suggestions on machine learning tasks, and the assistant suggested classification and regression. The user chose classification and specified that the target attribute would be the final grade. The assistant then asked for more information about the problem to help define it more precisely. The user decided to use 10,000 samples for the task\\* \hline
    9 & $utterance_3$, $context_3$, $PeTEL_3$ & Feeder & $PeTEL_3 :$
    
    \begin{lstlisting}[language=json]
{
    problem_type: classification, 
    target_variable: final grade, 
    features: None, 
    dataset_size: 10000, 
    performance_metrics: None, 
    validation_method: None, 
    classification_methods: None, 
    data_filters: [
        {column: final grade, condition: less than, value: passing grade}, 
        {column: None, condition: None, value: None}
        ], 
    business_goals: None, 
    additional_requirements: None, 
    model_preferences: None

}
    \end{lstlisting}
    
    \\* \hline
    10 & $PeTEL_3$ & PeTEL checker & $PeTEL\_is\_complete$ \\* \hline
    
    \multicolumn{4}{c}{Repeat step 6 - 10 until PeTEL is complete. Completed PeTEL is denoted by: $PeTEL^*$ } \\* \hline

    n & $PeTEL^*$, $context_n$ & PeTEL Descriptor & $Description: $ 
    
     The task is to predict the final\_grade of a student based on various features like study\_hours, attendance, participation, homework\_scores, and test\_scores. The dataset size is hypothetical. Performance is measured using accuracy, F1 score, and the confusion matrix. K-fold cross-validation is used for validation. Three classification methods are considered: random forest, SVM, and logistic regression. 
     
     The data is filtered to only include records where attendance is greater than 75 (assuming attendance is measured as a percentage) and study\_hours is more than 1. The business goal is to predict student performance to implement early interventions for those at risk of failing. Model interpretability is highlighted as an important additional requirement, and there's a preference for models with higher accuracy, but interpretability is also considered important.
    
    \\* \hline

    n+1 & $context_{n+1}$, $utterance_{n+1}$ & State Selector & $ds_{n+1} :$ Prediction Engineering \\* \hline

\caption{Interaction of different micro-agent in the Task Selection state. In the table, $utterance_1, utterance_2, utterance_3 $ are from Table~\ref{table:overall-cycle-classification}, and $summary, ML\_task$ are from Table~\ref{table:state-data-vis}.}
    \label{table:state-task-formulate}
\end{longtable}
}

\subsubsection{\textbf{Prediction Engineering State:}}

The third state, prediction engineering, is an integral part of the AutoML pipeline. This state takes the formulated task and prepares the dataset accordingly. The interaction between micro-agent agents in this state is essential as it directly influences the preparation of the dataset, which in turn impacts the accuracy and reliability of the predictions made. In this state, the micro-agent agents work collaboratively to interpret the task requirements, adjust the dataset to align with these requirements, and set the stage for accurate prediction generation. We will delve into the specifics of these interactions, referring to a table that outlines the interactions between the micro-agent agents during prediction engineering. This discussion will provide a comprehensive understanding of the role and effectiveness of the micro-agent agents in this crucial state of the AutoML pipeline.

{\small
\begin{longtable}{l|l|l|p{6cm}}\hline

    \hline {\textbf{\#}} & {\textbf{Input}} & {\textbf{Micro-Agent} } & {\textbf{Output}} \\ \hline 
    \endfirsthead
    
    \multicolumn{4}{c}%
    {{\bfseries \tablename\ \thetable{} -- continued from previous page}} \\
    \hline \multicolumn{1}{c|}{\textbf{\#}} & \multicolumn{1}{c|}{\textbf{Input}} & \multicolumn{1}{c}{\textbf{Micro-Agent}} & \multicolumn{1}{c}{\textbf{Output}} \\ \hline 
    \endhead
    
    \hline \multicolumn{4}{r}{{Continued on next page}} \\ \hline
    \endfoot

    \endlastfoot

    1 & $PeTEL_3$ & PeTEL to Attribute Converter & List of attributes \\* \hline

    2 & $PeTEL_3$ & Data Prepper & Prepares data with the conditions in PeTEL \\* \hline

    3 & $PeTEL_3$ & AutoML Interfacer & Calls the AutoML interface \\* \hline

\caption{Interaction of different micro-agent in the Prediction Engineering state.}
    \label{table:state-prediction-eng}
\end{longtable}
}

\subsubsection{\textbf{Model Training, Result Summary, and Recommendation State:}}

After the task is formulated, VIDS interfaces with AutoML tools (e.g. AutoSKLearn) and trains downstream model(s) based on the task formulation determined beforehand. As shown in Table~\ref{table:state-result}, from training performance of different models, VIDS generates summaries, including the results and recommendations based on user preferences defined in the task formulation. Our future work will be to interact with the user in this stage and evaluate different models based on the user's business goals.

{\small
\begin{longtable}{l|p{1.5cm}|l|p{6cm}}\hline

    \hline {\textbf{Step}} & {\textbf{Input}} & {\textbf{Micro-Agent} } & {\textbf{Output}} \\ \hline 
    \endfirsthead
    
    \multicolumn{4}{c}%
    {{\bfseries \tablename\ \thetable{} -- continued from previous page}} \\
    \hline \multicolumn{1}{c|}{\textbf{Step}} & \multicolumn{1}{c|}{\textbf{Input}} & \multicolumn{1}{c}{\textbf{Micro-Agent}} & \multicolumn{1}{c}{\textbf{Output}} \\ \hline 
    \endhead
    
    \hline \multicolumn{4}{r}{{Continued on next page}} \\ \hline
    \endfoot

    \endlastfoot

    1 & $context_n$, $PeTEL^*$, AutoML response & Result Summarizer & $Result :$ performance of each model based on evaluation criteria set in problem formulation.\\* \hline
    
    2 & $context_n$, $Result$ & Result Visualizer & $Output :$ Description of results in natural language.\\* \hline
    
\caption{Interaction of different micro-agent in the Task Formulation state. In the table, $utterance_1, utterance_2, utterance_3 $ are from Table~\ref{table:overall-cycle-classification}}
    \label{table:state-result}
\end{longtable}
}

\subsection{Prompt Engineering Taxonomy}
\label{subsec:taxonomy}

The successful collaboration between humans and artificial intelligence in complex tasks necessitates a comprehensive understanding of the various levels of interaction that occur between them. These levels span from Level 0, where AI is solely responsible for data processing, to Level 5, which involves the integration of evaluation criteria. 
Building upon the foundational work on taxonomy of prompt engineering (TELeR) by \citet{santu2023teler}, we put forward the notion of considering the depth of information that the System Role discloses to the Large Language Model (LLM). To illustrate, if a system role is well-delineated, it precludes its prompt from being classified as Level 0.
This study will specifically focus on three micro-agents: the Intent and State Detector, the Dialogue Summarizer, and the Conversation Manager.  Each of these micro-agents plays a unique and integral role in fostering a dynamic and functional dialogue between the user and the AI, leading to a more streamlined and efficient system overall. The revised taxonomy for these interaction levels is as follows:

\textbf{Level 0}: No directive is given. The focus is solely on the exchange of data.

\textbf{Level 1}: A simple one-sentence directive is provided, expressing the high-level goal of the task.

\textbf{Level 2}: A multi-sentence (paragraph-style) directive is given, expressing the high-level goal and the sub-tasks needed to achieve this goal.

\textbf{Level 3}: A complex directive is provided, expressing the high-level goal along with a bulleted list of subtasks that need to be performed.

\textbf{Level 4}: This level includes a complex directive that encompasses the following: 1) A description of the high-level goal, 2) A detailed bulleted list of subtasks, and 3) An explicit statement asking the LLM to explain its response.

\textbf{Level 5}: This level includes a complex directive that encompasses the following: 1) A description of the high-level goal, 2) A detailed bulleted list of subtasks, 3) An explicit statement asking the LLM to explain its response, and 4) A guideline about how the response should be evaluated.

By understanding these levels of interaction, we can maximize the potential benefits of AI and guide future research into user experience, system performance, and ethical considerations in AI applications.

\subsubsection{\textbf{Intent and State Detector micro-agent:}}
In terms of the taxonomy of prompts, the data for this micro-agent is as follows:

\begin{enumerate}
    \item \textbf{context:}  The user and the AI assistant discussed the summary of a dataset containing information about students in a school, including their demographic information, academic performance, and social life. The AI assistant suggested two suitable machine learning tasks based on the dataset: classification and regression. The user agreed that classification is a good choice to identify students who are at risk of failing and provide early intervention to improve their academic performance.
    \item \textbf{state:}  dataset\_understanding
    \item \textbf{utterance:}  Ok, from the description it seems like classification is a good choice.
\end{enumerate}













By following the prompt taxonomy we discussed earlier, Table~\ref{table:state-taxonomy} represents the response from the ChatGPT agent for each level of prompt. For this task, prompts of Level 4 and 5 are not applicable as the output should be strict JSON so that the output is used by other micro-agents.

{\small
\begin{longtable}{l|p{11cm}}\hline

    \hline {\textbf{Prompt Level}} & {\textbf{Output From ChatGPT}} \\ \hline 
    \endfirsthead
    
    \multicolumn{2}{c}%
    {{\bfseries \tablename\ \thetable{} -- continued from previous page}} \\
    \hline \multicolumn{1}{c|}{\textbf{Prompt Level}} & \multicolumn{1}{c}{\textbf{Output From ChatGPT}} \\ \hline 
    \endhead
    
    \hline \multicolumn{2}{r}{{Continued on next page}} \\ \hline
    \endfoot

    \endlastfoot

    0  & Can you tell me more about the dataset? What are the features and labels?\\* \hline
    
    1 & Intent: dataset\_understanding
    State: The user and the AI assistant have discussed a dataset containing information about students in a school, and the AI assistant has suggested two suitable machine learning tasks: classification and regression. The user has agreed that classification is a good choice to identify students who are at risk of failing and provide early intervention to improve their academic performance.\\* \hline

    2 &  \{ intent: Select problem, 
    current\_state: Dataset\_Understanding, 
    next\_state: Task\_Formulation \}\\* \hline

    3 &  \{ intent: Problem Execution, 
    current\_state: Dataset\_Understanding, 
    next\_state: Task\_Formulation \}\\* \hline


    
\caption{Intent and State Detector micro-agent response for different level of prompt engineering input.}
    \label{table:state-taxonomy}
\end{longtable}
}

\subsubsection{\textbf{Dialogue Summarizer micro-agent:}}

In terms of the taxonomy of prompts, the data for this micro-agent is:

\begin{itemize}
    \item {\textbf{history:} 
        \begin{enumerate}
            \item assistant: summary of Student Performance dataset
            \item user: From the summary you provided can you suggest me any machine learning task?
            \item assistant: $context_1$ from Table~\ref{table:state-task-formulate}
            \item user: Ok, from the description it seems like classification is a good choice.
        \end{enumerate}
    }

\end{itemize}
















By following the prompt taxonomy we discussed earlier, 
Table~\ref{table:dialog-sum-taxonomy} represents the response from the ChatGPT agent for each level of prompt. Similar to Intent and State Detector, prompts of Level 4 and 5 are not applicable for this task as the output is used by other micro-agents

{\small
\begin{longtable}{l|p{11cm}}\hline

    \hline {\textbf{Prompt Level}} & {\textbf{Output From ChatGPT}} \\ \hline 
    \endfirsthead
    
    \multicolumn{2}{c}%
    {{\bfseries \tablename\ \thetable{} -- continued from previous page}} \\
    \hline \multicolumn{1}{c|}{\textbf{Prompt Level}} & \multicolumn{1}{c}{\textbf{Output From ChatGPT}} \\ \hline 
    \endhead
    
    \hline \multicolumn{2}{r}{{Continued on next page}} \\ \hline
    \endfoot

    \endlastfoot

    0  & Great! Classification is a suitable choice for predicting whether a student will pass or fail based on their demographic information, academic performance, and social life. You can use various classification algorithms such as logistic regression, decision trees, random forests, and support vector machines to build a model that can accurately predict the target variable. Additionally, you can use techniques such as feature selection and hyperparameter tuning to improve the performance of the model.\\* \hline
    
    1 & Great! Classification is a suitable choice for predicting whether a student will pass or fail based on their demographic information, academic performance, and social life. This can help identify students who are at risk of failing and provide early intervention to improve their academic performance.\\* \hline

    2 & The user and assistant discussed the summary of a dataset containing information about students in a school, including their demographic information, academic performance, and social life. The assistant suggested two suitable machine learning tasks based on the dataset: classification and regression. The user decided that classification would be a good choice to identify students at risk of failing and provide early intervention.\\* \hline

    3 &  Description of results in natural language.\\* \hline
    
\caption{Dialog Summarizer micro-agent response for different level of prompt engineering input.}
    \label{table:dialog-sum-taxonomy}
\end{longtable}
}

\subsubsection{\textbf{Conversation Manager micro-agent}}

In terms of the taxonomy of prompts, the data for this micro-agent is as follows:

\begin{enumerate}
    \item \textbf{context:} The user and assistant discussed the dataset containing information about students in a school, including their demographic information, academic performance, and social life. The assistant suggested that the suitable machine learning tasks for this dataset are classification and regression, with classification being a good choice to identify students at risk of failing and provide early intervention to improve their academic performance. The user agreed that classification is a good choice.
    \item \textbf{state:} Dataset Visualization
    \item \textbf{utterance:} Ok, from the description it seems like classification is a good choice.
    \item \textbf{intent:} Select problem
    \item \textbf{mp\_resp:} \curlybr{\mquote{model}: \mquote{classification}, \mquote{reason}: \mquote{Classification is a good choice for identifying students at risk of failing and providing early intervention to improve their academic performance. The dataset contains demographic information, academic performance, and social life, which can be used as input features to predict the likelihood of a student failing or passing.}}
\end{enumerate}

















By following the prompt taxonomy we discussed earlier, 
Table~\ref{table:cm-taxonomy} represents the response from the ChatGPT agent for each level of prompt.

{\small
\begin{longtable}{l|p{11cm}}\hline

    \hline {\textbf{Prompt Level}} & {\textbf{Output From ChatGPT}} \\ \hline 
    \endfirsthead
    
    \multicolumn{2}{c}%
    {{\bfseries \tablename\ \thetable{} -- continued from previous page}} \\
    \hline \multicolumn{1}{c|}{\textbf{Prompt Level}} & \multicolumn{1}{c}{\textbf{Output From ChatGPT}} \\ \hline 
    \endhead
    
    \hline \multicolumn{2}{r}{{Continued on next page}} \\ \hline
    \endfoot

    \endlastfoot

    0  & The assistant then asked the user about the specific features in the dataset that could be used for classification. The user mentioned that features such as gender, age, ethnicity, parental education level, and previous academic performance could be used to predict whether a student is at risk of failing. The assistant agreed and suggested that the dataset could be split into training and testing sets to build and evaluate a classification model. The user agreed and asked the assistant for recommendations on which classification algorithms to use. The assistant suggested using logistic regression, decision trees, and random forests, as they are commonly used for classification tasks and can handle both categorical and numerical data. The user thanked the assistant for the suggestions and said they would explore these algorithms further.\\* \hline
    
    1 & Yes, based on the information you provided, classification seems like a suitable machine learning task for the dataset containing information about students in a school. It can help identify students who are at risk of failing and provide early intervention to improve their academic performance. Regression can also be a good choice to predict the academic performance of students based on their demographic and social information.
    \\* \hline

    2 &  Great! Now that we have agreed that classification is a suitable task for your dataset, let's move on to formulating the problem. Can you please provide me with more details about the specific problem you want to solve using this dataset? For example, do you want to predict which students are at risk of failing a particular subject or identify factors that contribute to academic success? This information will help us formulate a more precise ML task that meets your needs.\\* \hline

    3 &  N/A\\* \hline

    4 & Based on the given context, it seems like we have reached a consensus that classification is a suitable machine learning task for the dataset containing information about students in a school. Now, we need to formulate the problem to proceed further. 
    To do that, I suggest we move to the next conversation state, which is \mquote{problem\_formulation}. In this state, we can discuss the specific details of the classification task, such as the target variable, the features to be used, and the algorithm to be employed. 
    So, to move forward, can you please provide me with more information about the problem you want to solve using classification? This will help me formulate the problem and suggest relevant solutions to you.\\* \hline

    5 & Based on the current conversation state, it seems like we are in the Task\_Formulation state, and the user has agreed that classification is a good choice for the given dataset. Therefore, we can proceed with formulating the classification problem. 
    
    To do so, we need to determine the target variable that we want to predict. In this case, we want to identify students who are at risk of failing and provide early intervention to improve their academic performance. Therefore, we can use the \mquote{final grade} variable as our target variable, and classify students into two categories: those who are at risk of failing and those who are not.
    
    To achieve this, we can use a binary classification algorithm, such as logistic regression or decision tree. We can also use feature selection techniques to identify the most important features that contribute to the prediction of the target variable.
    
    Based on this, we can formulate the ML task as follows: Develop a binary classification model to predict whether a student is at risk of failing based on their demographic information, academic performance, and social life. The target variable is the final grade, and the model should be trained on a labeled dataset. Feature selection techniques can be used to identify the most important features.
    
    This task can be further refined based on the user\'s specific requirements and preferences. For example, the user may want to use a different classification algorithm or include additional features in the model.
    
    Evaluation criteria for the generated response:
    
    - The response provides a clear and concise formulation of the ML task based on the user\'s requirements and preferences.
    
    - The response takes into account the dataset description and the previous discussion between the user and the assistant.
    
    - The response suggests an appropriate classification algorithm and feature selection techniques based on the problem requirements.
    
    - The response is open-ended and allows for further refinement and customization based on the user\'s specific needs \\* \hline
    
\caption{Conversation Manager micro-agent response for different level of prompt engineering input.}
    \label{table:cm-taxonomy}
\end{longtable}
}

\section{Discussion}

As we navigate the complex task of automating task formulation and prediction engineering for an AutoML pipeline, ChatGPT emerges as a crucial ally. Its capabilities have streamlined this intricate process, paving the way for more accessible automation. Integral to this is the decomposition of larger tasks into smaller, specific micro-agents. This strategy enhances ChatGPT's performance and results in superior accuracy. We've observed that using targeted prompts, as opposed to a single overarching one, offers increased precision and control, leading to a more focused AI response and reducing the risk of errors and misunderstandings.

Progressing with ChatGPT as a cornerstone in our project, we find that it simplifies natural language processing and advances us towards our broader goal of democratizing Machine Learning. This pivotal role that ChatGPT plays fuels VIDS' capacity to tackle more nuanced and intricate tasks, guiding our trajectory for future endeavors.

Turning to our existing system, it is designed to engage with the user in active dialog for problem formulation and subsequently present the results in a user-friendly conversational format. However, at this stage, our focus remains on the front-end process, and we do not yet facilitate in-depth discussion or analysis of these results. Looking ahead, our vision for continuous improvement involves augmenting VIDS to assess the performance of various models based on the user's unique business requirements. This enhancement will elevate our capacity to cater to individual needs, improving user understanding and empowering more informed decision-making. This commitment to continuous evolution drives us closer to our ambition of democratizing Machine Learning.

\subsection{Fail cases}

When assessing the Intent and State Detection micro-agent, we confronted some area of failure in the ChatGPT model's performance. This issue manifested itself primarily in its inability to accurately decipher highly specific prompts, as described in Table~\ref{table:state-prompt} for the state detection task. Though the prompts distinctly defined both the current and subsequent states, ChatGPT consistently failed to correctly identify the intended state. One glaring example can be found in the user utterance, \mquote{Ok, from the description it seems like classification is a good choice}, found in the dataset descriptions in Table~\ref{table:state-data-vis}. Here, the user's clear intent to select a Machine Learning task (classification) should have led to the identification of `Task Selection' as the selected state. Yet, ChatGPT mistakenly attributed `Model Training' as the selected state. In an attempt to mitigate this failure, we introduced a modification to the prompt design to specify potential next states: \mquote{Next state should be from the following states - \curlybr{next\_states}}. In this case, \curlybr{next\_states} should have included [data\_visualization, task\_selection]. This remedial action has shown promise in enhancing the accuracy of the state selector.

Additionally, we encountered a significant number of failures during the development of the dialog summarization micro-agent. Specifically, ChatGPT exhibited a propensity to generate unrelated, or \mquote{hallucinated}, content when given few-shot learning examples. Our original process involved supplying a sample dialog between a user and an agent, along with its summary, in the expectation that ChatGPT would replicate this summarization approach. However, during the testing phase, it became evident that ChatGPT failed to understand the task correctly, treating the few-shot examples as part of the source text for summarization, rather than concentrating on the latest user input.

In conclusion, these cases represent significant challenges encountered in the development and testing phases of the ChatGPT model. Despite its advanced capabilities, the model displayed critical areas of failure in both the Intent and State Detection and dialog summarization micro-agents. Although we have introduced modifications to mitigate these issues and have seen some improvement, it is crucial to acknowledge these failures as opportunities for further research and development. The ability to accurately identify and rectify such errors is paramount in enhancing the model's robustness, efficiency, and overall performance. This analysis is instrumental in guiding our future efforts towards optimizing the ChatGPT model and bringing us closer to our ultimate goal of creating an AI that can effectively understand and engage with its users.

\section{Conclusion}

In this research, we have ventured into the realm of Large Language Models (LLMs) as personal data scientist (VIDS), with language acting as the pivotal interface linking LLMs and machine learning models. VIDS is architectured around four distinct dialogue states - Data Visualization, Task Formulation, Prediction Engineering, and Result Summary and Recommendation. Each of these states signifies a unique phase in the conversation and plays a substantial role in the overall user-system interaction.

We have introduced the concept of global micro-agents, which form an overarching structure, maintaining a cohesive narrative throughout the dialogue, irrespective of the specific state. Complementing these are the local micro-agents, which are integral to each state and play a crucial role in VIDS' functionality.

Despite the advanced capabilities of VIDS, it is crucial to acknowledge the areas of failure, particularly in the Intent and State Detection and dialog summarization micro-agents. While we have implemented modifications to mitigate these issues and have observed some improvements, these shortcomings highlight the need for further research and development. The identification and rectification of such errors are paramount in enhancing the model's robustness, efficiency, and overall performance.

In conclusion, this research serves as a significant milestone towards our ultimate goal of creating an AI data science assistant that can effectively understand and engage with its users. The insights gleaned from this study will steer our future efforts in optimizing the ChatGPT model, edging us closer to harnessing the full potential of AI in the field of data science. We are confident that the continued refinement of these models will pave the way for more intuitive and effective human-AI interactions, revolutionizing the way we approach complex tasks and data analysis.

\bibliography{llm}


\begin{thebibliography}{00}


\ifx \showCODEN    \undefined \def \showCODEN     #1{\unskip}     \fi
\ifx \showDOI      \undefined \def \showDOI       #1{{\tt DOI:}\penalty0{#1}\ }
  \fi
\ifx \showISBNx    \undefined \def \showISBNx     #1{\unskip}     \fi
\ifx \showISBNxiii \undefined \def \showISBNxiii  #1{\unskip}     \fi
\ifx \showISSN     \undefined \def \showISSN      #1{\unskip}     \fi
\ifx \showLCCN     \undefined \def \showLCCN      #1{\unskip}     \fi
\ifx \shownote     \undefined \def \shownote      #1{#1}          \fi
\ifx \showarticletitle \undefined \def \showarticletitle #1{#1}   \fi
\ifx \showURL      \undefined \def \showURL       #1{#1}          \fi

\bibitem[\protect\citeauthoryear{Aliannejadi, Chakraborty, R{\'{\i}}ssola, and
  Crestani}{Aliannejadi et~al\mbox{.}}{2020}]%
        {Eval-of-Multi-Turn-DS}
{Mohammad Aliannejadi}, {Manajit Chakraborty}, {Esteban~Andr{\'{e}}s
  R{\'{\i}}ssola}, {and} {Fabio Crestani}. 2020.
\newblock \showarticletitle{Harnessing Evolution of Multi-Turn Conversations
  for Effective Answer Retrieval}. In {\em {CHIIR} '20: Conference on Human
  Information Interaction and Retrieval, Vancouver, BC, Canada, March 14-18,
  2020}, {Heather~L. O'Brien}, {Luanne Freund}, {Ioannis Arapakis}, {Orland
  Hoeber}, {and} {Irene Lopatovska} (Eds.). {ACM}, 33--42.
\newblock
\showDOI{%
\url{http://dx.doi.org/10.1145/3343413.3377968}}


\bibitem[\protect\citeauthoryear{Baker, Gupta, Raskar, and Naik}{Baker
  et~al\mbox{.}}{2017}]%
        {baker2017accelerating}
{Bowen Baker}, {Otkrist Gupta}, {Ramesh Raskar}, {and} {Nikhil Naik}. 2017.
\newblock \showarticletitle{Accelerating neural architecture search using
  performance prediction}.
\newblock {\em arXiv preprint arXiv:1705.10823\/} (2017).
\newblock


\bibitem[\protect\citeauthoryear{Bengio}{Bengio}{2012}]%
        {bengio2012practical}
{Yoshua Bengio}. 2012.
\newblock \showarticletitle{Practical recommendations for gradient-based
  training of deep architectures}.
\newblock In {\em Neural networks: Tricks of the trade}. Springer, 437--478.
\newblock


\bibitem[\protect\citeauthoryear{Bergstra and Bengio}{Bergstra and
  Bengio}{2012}]%
        {bergstra2012random}
{James Bergstra} {and} {Yoshua Bengio}. 2012.
\newblock \showarticletitle{Random search for hyper-parameter optimization}.
\newblock {\em Journal of Machine Learning Research\/} {13}, Feb (2012),
  281--305.
\newblock


\bibitem[\protect\citeauthoryear{Bergstra, Yamins, and Cox}{Bergstra
  et~al\mbox{.}}{2013}]%
        {bergstra2013making}
{James Bergstra}, {Daniel Yamins}, {and} {David Cox}. 2013.
\newblock \showarticletitle{Making a science of model search: Hyperparameter
  optimization in hundreds of dimensions for vision architectures}. In {\em
  International conference on machine learning}. PMLR, 115--123.
\newblock


\bibitem[\protect\citeauthoryear{Bergstra, Bardenet, Bengio, and
  K{\'e}gl}{Bergstra et~al\mbox{.}}{2011}]%
        {bergstra2011algorithms}
{James~S Bergstra}, {R{\'e}mi Bardenet}, {Yoshua Bengio}, {and} {Bal{\'a}zs
  K{\'e}gl}. 2011.
\newblock \showarticletitle{Algorithms for hyper-parameter optimization}. In
  {\em Advances in neural information processing systems}. 2546--2554.
\newblock


\bibitem[\protect\citeauthoryear{Brown, Mann, Ryder, Subbiah, Kaplan, Dhariwal,
  Neelakantan, Shyam, Sastry, Askell, Agarwal, Herbert{-}Voss, Krueger,
  Henighan, Child, Ramesh, Ziegler, Wu, Winter, Hesse, Chen, Sigler, Litwin,
  Gray, Chess, Clark, Berner, McCandlish, Radford, Sutskever, and Amodei}{Brown
  et~al\mbox{.}}{2020}]%
        {gpt3}
{Tom~B. Brown}, {Benjamin Mann}, {Nick Ryder}, {Melanie Subbiah}, {Jared
  Kaplan}, {Prafulla Dhariwal}, {Arvind Neelakantan}, {Pranav Shyam}, {Girish
  Sastry}, {Amanda Askell}, {Sandhini Agarwal}, {Ariel Herbert{-}Voss},
  {Gretchen Krueger}, {Tom Henighan}, {Rewon Child}, {Aditya Ramesh},
  {Daniel~M. Ziegler}, {Jeffrey Wu}, {Clemens Winter}, {Christopher Hesse},
  {Mark Chen}, {Eric Sigler}, {Mateusz Litwin}, {Scott Gray}, {Benjamin Chess},
  {Jack Clark}, {Christopher Berner}, {Sam McCandlish}, {Alec Radford}, {Ilya
  Sutskever}, {and} {Dario Amodei}. 2020.
\newblock \showarticletitle{Language Models are Few-Shot Learners}. In {\em
  Advances in Neural Information Processing Systems 33: Annual Conference on
  Neural Information Processing Systems 2020, NeurIPS 2020, December 6-12,
  2020, virtual}, {Hugo Larochelle}, {Marc'Aurelio Ranzato}, {Raia Hadsell},
  {Maria{-}Florina Balcan}, {and} {Hsuan{-}Tien Lin} (Eds.).
\newblock
\showURL{%
\url{https://proceedings.neurips.cc/paper/2020/hash/1457c0d6bfcb4967418bfb8ac142f64a-Abstract.html}}


\bibitem[\protect\citeauthoryear{Campagna, Foryciarz, Moradshahi, and
  Lam}{Campagna et~al\mbox{.}}{2020}]%
        {DBLP:conf/acl/CampagnaFML20}
{Giovanni Campagna}, {Agata Foryciarz}, {Mehrad Moradshahi}, {and} {Monica~S.
  Lam}. 2020.
\newblock \showarticletitle{Zero-Shot Transfer Learning with Synthesized Data
  for Multi-Domain Dialogue State Tracking}. In {\em Proceedings of the 58th
  Annual Meeting of the Association for Computational Linguistics, {ACL} 2020,
  Online, July 5-10, 2020}, {Dan Jurafsky}, {Joyce Chai}, {Natalie Schluter},
  {and} {Joel~R. Tetreault} (Eds.). Association for Computational Linguistics,
  122--132.
\newblock
\showDOI{%
\url{http://dx.doi.org/10.18653/v1/2020.acl-main.12}}


\bibitem[\protect\citeauthoryear{Chen, Lv, Wang, Zhu, Tan, and Yu}{Chen
  et~al\mbox{.}}{2020}]%
        {DBLP:conf/aaai/0002LWZT020}
{Lu Chen}, {Boer Lv}, {Chi Wang}, {Su Zhu}, {Bowen Tan}, {and} {Kai Yu}. 2020.
\newblock \showarticletitle{Schema-Guided Multi-Domain Dialogue State Tracking
  with Graph Attention Neural Networks}. In {\em The Thirty-Fourth {AAAI}
  Conference on Artificial Intelligence, {AAAI} 2020, The Thirty-Second
  Innovative Applications of Artificial Intelligence Conference, {IAAI} 2020,
  The Tenth {AAAI} Symposium on Educational Advances in Artificial
  Intelligence, {EAAI} 2020, New York, NY, USA, February 7-12, 2020}. {AAAI}
  Press, 7521--7528.
\newblock
\showURL{%
\url{https://aaai.org/ojs/index.php/AAAI/article/view/6250}}


\bibitem[\protect\citeauthoryear{Chen, Zaharia, and Zou}{Chen
  et~al\mbox{.}}{2023}]%
        {FrugalGPT-red-cost}
{Lingjiao Chen}, {Matei Zaharia}, {and} {James Zou}. 2023.
\newblock \showarticletitle{FrugalGPT: How to Use Large Language Models While
  Reducing Cost and Improving Performance}.
\newblock {\em CoRR\/}  {abs/2305.05176} (2023).
\newblock
\showDOI{%
\url{http://dx.doi.org/10.48550/arXiv.2305.05176}}


\bibitem[\protect\citeauthoryear{Chowdhery, Narang, Devlin, Bosma, Mishra,
  Roberts, Barham, Chung, Sutton, Gehrmann, Schuh, Shi, Tsvyashchenko, Maynez,
  Rao, Barnes, Tay, Shazeer, Prabhakaran, Reif, Du, Hutchinson, Pope, Bradbury,
  Austin, Isard, Gur{-}Ari, Yin, Duke, Levskaya, Ghemawat, Dev, Michalewski,
  Garcia, Misra, Robinson, Fedus, Zhou, Ippolito, Luan, Lim, Zoph, Spiridonov,
  Sepassi, Dohan, Agrawal, Omernick, Dai, Pillai, Pellat, Lewkowycz, Moreira,
  Child, Polozov, Lee, Zhou, Wang, Saeta, Diaz, Firat, Catasta, Wei,
  Meier{-}Hellstern, Eck, Dean, Petrov, and Fiedel}{Chowdhery
  et~al\mbox{.}}{2022}]%
        {palm}
{Aakanksha Chowdhery}, {Sharan Narang}, {Jacob Devlin}, {Maarten Bosma},
  {Gaurav Mishra}, {Adam Roberts}, {Paul Barham}, {Hyung~Won Chung}, {Charles
  Sutton}, {Sebastian Gehrmann}, {Parker Schuh}, {Kensen Shi}, {Sasha
  Tsvyashchenko}, {Joshua Maynez}, {Abhishek Rao}, {Parker Barnes}, {Yi Tay},
  {Noam Shazeer}, {Vinodkumar Prabhakaran}, {Emily Reif}, {Nan Du}, {Ben
  Hutchinson}, {Reiner Pope}, {James Bradbury}, {Jacob Austin}, {Michael
  Isard}, {Guy Gur{-}Ari}, {Pengcheng Yin}, {Toju Duke}, {Anselm Levskaya},
  {Sanjay Ghemawat}, {Sunipa Dev}, {Henryk Michalewski}, {Xavier Garcia},
  {Vedant Misra}, {Kevin Robinson}, {Liam Fedus}, {Denny Zhou}, {Daphne
  Ippolito}, {David Luan}, {Hyeontaek Lim}, {Barret Zoph}, {Alexander
  Spiridonov}, {Ryan Sepassi}, {David Dohan}, {Shivani Agrawal}, {Mark
  Omernick}, {Andrew~M. Dai}, {Thanumalayan~Sankaranarayana Pillai}, {Marie
  Pellat}, {Aitor Lewkowycz}, {Erica Moreira}, {Rewon Child}, {Oleksandr
  Polozov}, {Katherine Lee}, {Zongwei Zhou}, {Xuezhi Wang}, {Brennan Saeta},
  {Mark Diaz}, {Orhan Firat}, {Michele Catasta}, {Jason Wei}, {Kathy
  Meier{-}Hellstern}, {Douglas Eck}, {Jeff Dean}, {Slav Petrov}, {and} {Noah
  Fiedel}. 2022.
\newblock \showarticletitle{PaLM: Scaling Language Modeling with Pathways}.
\newblock {\em CoRR\/}  {abs/2204.02311} (2022).
\newblock
\showDOI{%
\url{http://dx.doi.org/10.48550/arXiv.2204.02311}}


\bibitem[\protect\citeauthoryear{Chu, Ilyas, Krishnan, and Wang}{Chu
  et~al\mbox{.}}{2016}]%
        {chu2016data}
{Xu Chu}, {Ihab~F Ilyas}, {Sanjay Krishnan}, {and} {Jiannan Wang}. 2016.
\newblock \showarticletitle{Data cleaning: Overview and emerging challenges}.
  In {\em Proceedings of the 2016 international conference on Management of
  Data}. ACM, 2201--2206.
\newblock


\bibitem[\protect\citeauthoryear{Feurer, Klein, Eggensperger, Springenberg,
  Blum, and Hutter}{Feurer et~al\mbox{.}}{2015}]%
        {feurer2015efficient}
{Matthias Feurer}, {Aaron Klein}, {Katharina Eggensperger}, {Jost
  Springenberg}, {Manuel Blum}, {and} {Frank Hutter}. 2015.
\newblock \showarticletitle{Efficient and robust automated machine learning}.
  In {\em Advances in Neural Information Processing Systems}. 2962--2970.
\newblock


\bibitem[\protect\citeauthoryear{Henderson, Thomson, and Williams}{Henderson
  et~al\mbox{.}}{2014a}]%
        {henderson2014second}
{Matthew Henderson}, {Blaise Thomson}, {and} {Jason~D Williams}. 2014a.
\newblock \showarticletitle{The second dialog state tracking challenge}. In
  {\em Proceedings of the 15th annual meeting of the special interest group on
  discourse and dialogue (SIGDIAL)}. 263--272.
\newblock


\bibitem[\protect\citeauthoryear{Henderson, Thomson, and Williams}{Henderson
  et~al\mbox{.}}{2014b}]%
        {henderson2014third}
{Matthew Henderson}, {Blaise Thomson}, {and} {Jason~D. Williams}. 2014b.
\newblock \showarticletitle{The third Dialog State Tracking Challenge}. In {\em
  2014 IEEE Spoken Language Technology Workshop (SLT)}. 324--329.
\newblock
\showDOI{%
\url{http://dx.doi.org/10.1109/SLT.2014.7078595}}


\bibitem[\protect\citeauthoryear{Hutter, Hoos, and Leyton-Brown}{Hutter
  et~al\mbox{.}}{2011}]%
        {hutter2011sequential}
{Frank Hutter}, {Holger~H Hoos}, {and} {Kevin Leyton-Brown}. 2011.
\newblock \showarticletitle{Sequential model-based optimization for general
  algorithm configuration}. In {\em International Conference on Learning and
  Intelligent Optimization}. Springer, 507--523.
\newblock


\bibitem[\protect\citeauthoryear{Ilyas, Chu, et~al\mbox{.}}{Ilyas
  et~al\mbox{.}}{2015}]%
        {ilyas2015trends}
{Ihab~F Ilyas}, {Xu Chu}, {and} {others}. 2015.
\newblock \showarticletitle{Trends in cleaning relational data: Consistency and
  deduplication}.
\newblock {\em Foundations and Trends in Databases\/} {5}, 4 (2015), 281--393.
\newblock


\bibitem[\protect\citeauthoryear{Kanter and Veeramachaneni}{Kanter and
  Veeramachaneni}{2015}]%
        {kanter2015deep}
{James~Max Kanter} {and} {Kalyan Veeramachaneni}. 2015.
\newblock \showarticletitle{Deep feature synthesis: Towards automating data
  science endeavors}. In {\em Data Science and Advanced Analytics (DSAA), 2015.
  36678 2015. IEEE International Conference on}. IEEE, 1--10.
\newblock


\bibitem[\protect\citeauthoryear{Kapelonis, Georgiou, and Potamianos}{Kapelonis
  et~al\mbox{.}}{2022}]%
        {kapelonis_multi-task_2022}
{Eleftherios Kapelonis}, {Efthymios Georgiou}, {and} {Alexandros Potamianos}.
  2022.
\newblock A Multi-Task {BERT} Model for Schema-Guided Dialogue State Tracking.
\newblock   (2022).
\newblock
\showDOI{%
\url{http://dx.doi.org/10.48550/arXiv.2207.00828}}


\bibitem[\protect\citeauthoryear{Karmaker, Hassan, Smith, Xu, Zhai, and
  Veeramachaneni}{Karmaker et~al\mbox{.}}{2021}]%
        {karmaker2021automl}
{Shubhra~Kanti Karmaker}, {Md~Mahadi Hassan}, {Micah~J Smith}, {Lei Xu},
  {Chengxiang Zhai}, {and} {Kalyan Veeramachaneni}. 2021.
\newblock \showarticletitle{AutoML to Date and Beyond: Challenges and
  Opportunities}.
\newblock {\em ACM Computing Surveys (CSUR)\/} {54}, 8 (2021), 1--36.
\newblock


\bibitem[\protect\citeauthoryear{Katz, Shin, and Song}{Katz
  et~al\mbox{.}}{2016}]%
        {katz2016explorekit}
{Gilad Katz}, {Eui Chul~Richard Shin}, {and} {Dawn Song}. 2016.
\newblock \showarticletitle{Explorekit: Automatic feature generation and
  selection}. In {\em Data Mining (ICDM), 2016 IEEE 16th International
  Conference on}. IEEE, 979--984.
\newblock


\bibitem[\protect\citeauthoryear{Kaul, Maheshwary, and Pudi}{Kaul
  et~al\mbox{.}}{2017}]%
        {kaul2017autolearn}
{Ambika Kaul}, {Saket Maheshwary}, {and} {Vikram Pudi}. 2017.
\newblock \showarticletitle{AutoLearn-Automated Feature Generation and
  Selection}. In {\em Data Mining (ICDM), 2017 IEEE International Conference
  on}. IEEE, 217--226.
\newblock


\bibitem[\protect\citeauthoryear{Khatri, Goel, Hedayatnia, Metanillou,
  Venkatesh, Gabriel, and Mandal}{Khatri et~al\mbox{.}}{2018}]%
        {dialog-topic-modeling}
{Chandra Khatri}, {Rahul Goel}, {Behnam Hedayatnia}, {Angeliki Metanillou},
  {Anushree Venkatesh}, {Raefer Gabriel}, {and} {Arindam Mandal}. 2018.
\newblock \showarticletitle{Contextual Topic Modeling For Dialog Systems}. In
  {\em 2018 IEEE Spoken Language Technology Workshop (SLT)}. 892--899.
\newblock
\showDOI{%
\url{http://dx.doi.org/10.1109/SLT.2018.8639552}}


\bibitem[\protect\citeauthoryear{Khurana, Samulowitz, and Turaga}{Khurana
  et~al\mbox{.}}{2017}]%
        {khurana2017feature}
{Udayan Khurana}, {Horst Samulowitz}, {and} {Deepak Turaga}. 2017.
\newblock \showarticletitle{Feature Engineering for Predictive Modeling using
  Reinforcement Learning}.
\newblock {\em arXiv preprint arXiv:1709.07150\/} (2017).
\newblock


\bibitem[\protect\citeauthoryear{Liao, Zhu, Long, and Chua}{Liao
  et~al\mbox{.}}{2021}]%
        {dst-recursive-inference}
{Lizi Liao}, {Tongyao Zhu}, {Le~Hong Long}, {and} {Tat{-}Seng Chua}. 2021.
\newblock \showarticletitle{Multi-domain Dialogue State Tracking with Recursive
  Inference}. In {\em {WWW} '21: The Web Conference 2021, Virtual Event /
  Ljubljana, Slovenia, April 19-23, 2021}, {Jure Leskovec}, {Marko Grobelnik},
  {Marc Najork}, {Jie Tang}, {and} {Leila Zia} (Eds.). {ACM} / {IW3C2},
  2568--2577.
\newblock
\showDOI{%
\url{http://dx.doi.org/10.1145/3442381.3450134}}


\bibitem[\protect\citeauthoryear{Liu, Zoph, Shlens, Hua, Li, Fei-Fei, Yuille,
  Huang, and Murphy}{Liu et~al\mbox{.}}{2017b}]%
        {liu2017progressive}
{Chenxi Liu}, {Barret Zoph}, {Jonathon Shlens}, {Wei Hua}, {Li-Jia Li}, {Li
  Fei-Fei}, {Alan Yuille}, {Jonathan Huang}, {and} {Kevin Murphy}. 2017b.
\newblock \showarticletitle{Progressive neural architecture search}.
\newblock {\em arXiv preprint arXiv:1712.00559\/} (2017).
\newblock


\bibitem[\protect\citeauthoryear{Liu, Simonyan, Vinyals, Fernando, and
  Kavukcuoglu}{Liu et~al\mbox{.}}{2017a}]%
        {liu2017hierarchical}
{Hanxiao Liu}, {Karen Simonyan}, {Oriol Vinyals}, {Chrisantha Fernando}, {and}
  {Koray Kavukcuoglu}. 2017a.
\newblock \showarticletitle{Hierarchical representations for efficient
  architecture search}.
\newblock {\em arXiv preprint arXiv:1711.00436\/} (2017).
\newblock


\bibitem[\protect\citeauthoryear{Maclaurin, Duvenaud, and Adams}{Maclaurin
  et~al\mbox{.}}{2015}]%
        {maclaurin2015gradient}
{Dougal Maclaurin}, {David Duvenaud}, {and} {Ryan Adams}. 2015.
\newblock \showarticletitle{Gradient-based hyperparameter optimization through
  reversible learning}. In {\em International Conference on Machine Learning}.
  2113--2122.
\newblock


\bibitem[\protect\citeauthoryear{Mountantonakis and Tzitzikas}{Mountantonakis
  and Tzitzikas}{2017}]%
        {mountantonakis2017linked}
{Michalis Mountantonakis} {and} {Yannis Tzitzikas}. 2017.
\newblock \showarticletitle{How linked data can aid machine learning-based
  tasks}. In {\em International Conference on Theory and Practice of Digital
  Libraries}. Springer, 155--168.
\newblock


\bibitem[\protect\citeauthoryear{Olabiyi, Bhattarai, Bruss, and Kulis}{Olabiyi
  et~al\mbox{.}}{2020}]%
        {DLGNet-task}
{Oluwatobi~O. Olabiyi}, {Prarthana Bhattarai}, {C.~Bayan Bruss}, {and} {Zachary
  Kulis}. 2020.
\newblock \showarticletitle{DLGNet-Task: An End-to-end Neural Network Framework
  for Modeling Multi-turn Multi-domain Task-Oriented Dialogue}.
\newblock {\em CoRR\/}  {abs/2010.01693} (2020).
\newblock
\showURL{%
\url{https://arxiv.org/abs/2010.01693}}


\bibitem[\protect\citeauthoryear{Ouyang, Wu, Jiang, Almeida, Wainwright,
  Mishkin, Zhang, Agarwal, Slama, Ray, Schulman, Hilton, Kelton, Miller,
  Simens, Askell, Welinder, Christiano, Leike, and Lowe}{Ouyang
  et~al\mbox{.}}{2022}]%
        {RLHF}
{Long Ouyang}, {Jeff Wu}, {Xu Jiang}, {Diogo Almeida}, {Carroll~L. Wainwright},
  {Pamela Mishkin}, {Chong Zhang}, {Sandhini Agarwal}, {Katarina Slama}, {Alex
  Ray}, {John Schulman}, {Jacob Hilton}, {Fraser Kelton}, {Luke Miller},
  {Maddie Simens}, {Amanda Askell}, {Peter Welinder}, {Paul~F. Christiano},
  {Jan Leike}, {and} {Ryan Lowe}. 2022.
\newblock \showarticletitle{Training language models to follow instructions
  with human feedback}.
\newblock {\em CoRR\/}  {abs/2203.02155} (2022).
\newblock
\showDOI{%
\url{http://dx.doi.org/10.48550/arXiv.2203.02155}}


\bibitem[\protect\citeauthoryear{Pedregosa, Varoquaux, Gramfort, Michel,
  Thirion, Grisel, Blondel, Prettenhofer, Weiss, Dubourg, Vanderplas, Passos,
  Cournapeau, Brucher, Perrot, and Duchesnay}{Pedregosa et~al\mbox{.}}{2011}]%
        {scikit-learn}
{F. Pedregosa}, {G. Varoquaux}, {A. Gramfort}, {V. Michel}, {B. Thirion}, {O.
  Grisel}, {M. Blondel}, {P. Prettenhofer}, {R. Weiss}, {V. Dubourg}, {J.
  Vanderplas}, {A. Passos}, {D. Cournapeau}, {M. Brucher}, {M. Perrot}, {and}
  {E. Duchesnay}. 2011.
\newblock \showarticletitle{Scikit-learn: Machine Learning in {P}ython}.
\newblock {\em Journal of Machine Learning Research\/}  {12} (2011),
  2825--2830.
\newblock


\bibitem[\protect\citeauthoryear{Peng, Galley, He, Brockett, Liden, Nouri, Yu,
  Dolan, and Gao}{Peng et~al\mbox{.}}{2022}]%
        {godel}
{Baolin Peng}, {Michel Galley}, {Pengcheng He}, {Chris Brockett}, {Lars Liden},
  {Elnaz Nouri}, {Zhou Yu}, {Bill Dolan}, {and} {Jianfeng Gao}. 2022.
\newblock GODEL: Large-Scale Pre-training for Goal-Directed Dialog.
\newblock arXiv.   (June 2022).
\newblock
\showURL{%
\url{https://www.microsoft.com/en-us/research/publication/godel-large-scale-pre-training-for-goal-directed-dialog/}}


\bibitem[\protect\citeauthoryear{Pham, Guan, Zoph, Le, and Dean}{Pham
  et~al\mbox{.}}{2018}]%
        {pham2018efficient}
{Hieu Pham}, {Melody~Y Guan}, {Barret Zoph}, {Quoc~V Le}, {and} {Jeff Dean}.
  2018.
\newblock \showarticletitle{Efficient Neural Architecture Search via Parameter
  Sharing}.
\newblock {\em arXiv preprint arXiv:1802.03268\/} (2018).
\newblock


\bibitem[\protect\citeauthoryear{Rabinovich, Vetzler, Boaz, Kumar, Pandey, and
  Anaby-Tavor}{Rabinovich et~al\mbox{.}}{2022}]%
        {rabinovich_gaining_2022}
{Ella Rabinovich}, {Matan Vetzler}, {David Boaz}, {Vineet Kumar}, {Gaurav
  Pandey}, {and} {Ateret Anaby-Tavor}. 2022.
\newblock Gaining Insights into Unrecognized User Utterances in Task-Oriented
  Dialog Systems.
\newblock   (2022).
\newblock
\showURL{%
\url{http://arxiv.org/abs/2204.05158}}


\bibitem[\protect\citeauthoryear{Real, Aggarwal, Huang, and Le}{Real
  et~al\mbox{.}}{2018}]%
        {real2018regularized}
{Esteban Real}, {Alok Aggarwal}, {Yanping Huang}, {and} {Quoc~V Le}. 2018.
\newblock \showarticletitle{Regularized evolution for image classifier
  architecture search}.
\newblock {\em arXiv preprint arXiv:1802.01548\/} (2018).
\newblock


\bibitem[\protect\citeauthoryear{Romero, Wang, Zimmerman, Steinfeld, and
  Tomasic}{Romero et~al\mbox{.}}{2021}]%
        {task-oriented-ds}
{Oscar~J. Romero}, {Antian Wang}, {John Zimmerman}, {Aaron Steinfeld}, {and}
  {Anthony Tomasic}. 2021.
\newblock \showarticletitle{A Task-Oriented Dialogue Architecture via
  Transformer Neural Language Models and Symbolic Injection}. In {\em
  Proceedings of the 22nd Annual Meeting of the Special Interest Group on
  Discourse and Dialogue, SIGdial 2021, Singapore and Online, July 29-31,
  2021}, {Haizhou Li}, {Gina{-}Anne Levow}, {Zhou Yu}, {Chitralekha Gupta},
  {Berrak Sisman}, {Siqi Cai}, {David Vandyke}, {Nina Dethlefs}, {Yan Wu},
  {and} {Junyi~Jessy Li} (Eds.). Association for Computational Linguistics,
  438--444.
\newblock
\showURL{%
\url{https://aclanthology.org/2021.sigdial-1.46}}


\bibitem[\protect\citeauthoryear{Santu and Feng}{Santu and Feng}{2023}]%
        {santu2023teler}
{Shubhra Kanti~Karmaker Santu} {and} {Dongji Feng}. 2023.
\newblock TELeR: A General Taxonomy of LLM Prompts for Benchmarking Complex
  Tasks.
\newblock   (2023).
\newblock


\bibitem[\protect\citeauthoryear{Sarkar, Bijoy, Saba, Feng, Mahajan, Amin,
  Islam, and Karmaker}{Sarkar et~al\mbox{.}}{2023}]%
        {sarkar2023ad}
{Souvika Sarkar}, {Biddut~Sarker Bijoy}, {Syeda~Jannatus Saba}, {Dongji Feng},
  {Yash Mahajan}, {Mohammad~Ruhul Amin}, {Sheikh~Rabiul Islam}, {and}
  {Shubhra~Kanti Karmaker}. 2023.
\newblock \showarticletitle{Ad-Hoc Monitoring of COVID-19 Global Research
  Trends for Well-Informed Policy Making}.
\newblock {\em ACM Transactions on Intelligent Systems and Technology\/} {14},
  2 (2023), 1--28.
\newblock


\bibitem[\protect\citeauthoryear{Sarkar, Feng, and Santu}{Sarkar
  et~al\mbox{.}}{2022}]%
        {sarkar2022exploring}
{Souvika Sarkar}, {Dongji Feng}, {and} {Shubhra Kanti~Karmaker Santu}. 2022.
\newblock \showarticletitle{Exploring Universal Sentence Encoders for Zero-shot
  Text Classification}. In {\em Proceedings of the 2nd Conference of the
  Asia-Pacific Chapter of the Association for Computational Linguistics and the
  12th International Joint Conference on Natural Language Processing}.
  135--147.
\newblock


\bibitem[\protect\citeauthoryear{Shen, Song, Tan, Li, Lu, and Zhuang}{Shen
  et~al\mbox{.}}{2023}]%
        {HuggingGPT}
{Yongliang Shen}, {Kaitao Song}, {Xu Tan}, {Dongsheng Li}, {Weiming Lu}, {and}
  {Yueting Zhuang}. 2023.
\newblock \showarticletitle{HuggingGPT: Solving {AI} Tasks with ChatGPT and its
  Friends in HuggingFace}.
\newblock {\em CoRR\/}  {abs/2303.17580} (2023).
\newblock
\showDOI{%
\url{http://dx.doi.org/10.48550/arXiv.2303.17580}}


\bibitem[\protect\citeauthoryear{Snoek, Larochelle, and Adams}{Snoek
  et~al\mbox{.}}{2012}]%
        {snoek2012practical}
{Jasper Snoek}, {Hugo Larochelle}, {and} {Ryan~P Adams}. 2012.
\newblock \showarticletitle{Practical bayesian optimization of machine learning
  algorithms}. In {\em Advances in neural information processing systems}.
  2951--2959.
\newblock


\bibitem[\protect\citeauthoryear{Swearingen, Drevo, Cyphers, Cuesta-Infante,
  Ross, and Veeramachaneni}{Swearingen et~al\mbox{.}}{2017}]%
        {swearingen2017atm}
{Thomas Swearingen}, {Will Drevo}, {Bennett Cyphers}, {Alfredo Cuesta-Infante},
  {Arun Ross}, {and} {Kalyan Veeramachaneni}. 2017.
\newblock \showarticletitle{ATM: A distributed, collaborative, scalable system
  for automated machine learning}. In {\em 2017 IEEE International Conference
  on Big Data (Big Data)}. IEEE, 151--162.
\newblock


\bibitem[\protect\citeauthoryear{Swersky, Snoek, and Adams}{Swersky
  et~al\mbox{.}}{2013}]%
        {swersky2013multi}
{Kevin Swersky}, {Jasper Snoek}, {and} {Ryan~P Adams}. 2013.
\newblock \showarticletitle{Multi-task bayesian optimization}. In {\em Advances
  in neural information processing systems}. 2004--2012.
\newblock


\bibitem[\protect\citeauthoryear{Thornton, Hutter, Hoos, and
  Leyton-Brown}{Thornton et~al\mbox{.}}{2013}]%
        {thornton2013auto}
{Chris Thornton}, {Frank Hutter}, {Holger~H Hoos}, {and} {Kevin Leyton-Brown}.
  2013.
\newblock \showarticletitle{Auto-WEKA: Combined selection and hyperparameter
  optimization of classification algorithms}. In {\em Proceedings of the 19th
  ACM SIGKDD international conference on Knowledge discovery and data mining}.
  ACM, 847--855.
\newblock


\bibitem[\protect\citeauthoryear{Touvron, Lavril, Izacard, Martinet, Lachaux,
  Lacroix, Rozi{\`{e}}re, Goyal, Hambro, Azhar, Rodriguez, Joulin, Grave, and
  Lample}{Touvron et~al\mbox{.}}{2023}]%
        {LLaMA}
{Hugo Touvron}, {Thibaut Lavril}, {Gautier Izacard}, {Xavier Martinet},
  {Marie{-}Anne Lachaux}, {Timoth{\'{e}}e Lacroix}, {Baptiste Rozi{\`{e}}re},
  {Naman Goyal}, {Eric Hambro}, {Faisal Azhar}, {Aur{\'{e}}lien Rodriguez},
  {Armand Joulin}, {Edouard Grave}, {and} {Guillaume Lample}. 2023.
\newblock \showarticletitle{LLaMA: Open and Efficient Foundation Language
  Models}.
\newblock {\em CoRR\/}  {abs/2302.13971} (2023).
\newblock
\showDOI{%
\url{http://dx.doi.org/10.48550/arXiv.2302.13971}}


\bibitem[\protect\citeauthoryear{van~den Bosch}{van~den Bosch}{2017}]%
        {van2017automatic}
{Suzanne van~den Bosch}. 2017.
\newblock \showarticletitle{Automatic feature generation and selection in
  predictive analytics solutions}.
\newblock {\em Master's thesis, Faculty of Science, Radboud University\/} {3},
  1 (2017), 3--1.
\newblock


\bibitem[\protect\citeauthoryear{Veron, Galibert, Bernard, and Rosset}{Veron
  et~al\mbox{.}}{}]%
        {veron_attention_2022}
{Mathilde Veron}, {Olivier Galibert}, {Guillaume Bernard}, {and} {Sophie
  Rosset}.
\newblock \showarticletitle{Attention Modulation for Zero-Shot Cross-Domain
  Dialogue State Tracking}. In {\em Proceedings of the 3rd Workshop on
  Computational Approaches to Discourse} (2022-10). International Conference on
  Computational Linguistics, 86--91.
\newblock
\showURL{%
\url{https://aclanthology.org/2022.codi-1.11}}


\bibitem[\protect\citeauthoryear{Vlasov, Mosig, and Nichol}{Vlasov
  et~al\mbox{.}}{2020}]%
        {vlasov2020dialogue}
{Vladimir Vlasov}, {Johannes E.~M. Mosig}, {and} {Alan Nichol}. 2020.
\newblock Dialogue Transformers.
\newblock   (2020).
\newblock


\bibitem[\protect\citeauthoryear{Wang, Guo, and Zhu}{Wang
  et~al\mbox{.}}{2020}]%
        {DBLP:conf/emnlp/WangGZ20}
{Yexiang Wang}, {Yi Guo}, {and} {Siqi Zhu}. 2020.
\newblock \showarticletitle{Slot Attention with Value Normalization for
  Multi-Domain Dialogue State Tracking}. In {\em Proceedings of the 2020
  Conference on Empirical Methods in Natural Language Processing, {EMNLP} 2020,
  Online, November 16-20, 2020}, {Bonnie Webber}, {Trevor Cohn}, {Yulan He},
  {and} {Yang Liu} (Eds.). Association for Computational Linguistics,
  3019--3028.
\newblock
\showDOI{%
\url{http://dx.doi.org/10.18653/v1/2020.emnlp-main.243}}


\bibitem[\protect\citeauthoryear{Wang, Seyler, Santu, and Zhai}{Wang
  et~al\mbox{.}}{2017}]%
        {wang2017study}
{Yiren Wang}, {Dominic Seyler}, {Shubhra Kanti~Karmaker Santu}, {and}
  {ChengXiang Zhai}. 2017.
\newblock \showarticletitle{A study of feature construction for text-based
  forecasting of time series variables}. In {\em Proceedings of the 2017 ACM on
  Conference on Information and Knowledge Management}. 2347--2350.
\newblock


\bibitem[\protect\citeauthoryear{Williams}{Williams}{2013}]%
        {DBLP:conf/sigdial/Williams13}
{Jason Williams}. 2013.
\newblock \showarticletitle{Multi-domain learning and generalization in dialog
  state tracking}. In {\em Proceedings of the {SIGDIAL} 2013 Conference, The
  14th Annual Meeting of the Special Interest Group on Discourse and Dialogue,
  22-24 August 2013, SUPELEC, Metz, France}. The Association for Computer
  Linguistics, 433--441.
\newblock
\showURL{%
\url{https://aclanthology.org/W13-4068/}}


\bibitem[\protect\citeauthoryear{Witten and Frank}{Witten and Frank}{2002}]%
        {witten2002data}
{Ian~H Witten} {and} {Eibe Frank}. 2002.
\newblock \showarticletitle{Data mining: practical machine learning tools and
  techniques with Java implementations}.
\newblock {\em Acm Sigmod Record\/} {31}, 1 (2002), 76--77.
\newblock


\bibitem[\protect\citeauthoryear{Wu, Zhang, Li, and Yu}{Wu
  et~al\mbox{.}}{2021}]%
        {Dialog-model-finetuning}
{Qingyang Wu}, {Yichi Zhang}, {Yu Li}, {and} {Zhou Yu}. 2021.
\newblock \showarticletitle{Alternating Recurrent Dialog Model with Large-scale
  Pre-trained Language Models}. In {\em Proceedings of the 16th Conference of
  the European Chapter of the Association for Computational Linguistics: Main
  Volume, {EACL} 2021, Online, April 19 - 23, 2021}, {Paola Merlo}, {J{\"{o}}rg
  Tiedemann}, {and} {Reut Tsarfaty} (Eds.). Association for Computational
  Linguistics, 1292--1301.
\newblock
\showDOI{%
\url{http://dx.doi.org/10.18653/v1/2021.eacl-main.110}}


\bibitem[\protect\citeauthoryear{Xu, Szlam, and Weston}{Xu
  et~al\mbox{.}}{2022}]%
        {beyond-goldfish-memory}
{Jing Xu}, {Arthur Szlam}, {and} {Jason Weston}. 2022.
\newblock \showarticletitle{Beyond Goldfish Memory: Long-Term Open-Domain
  Conversation}. In {\em Proceedings of the 60th Annual Meeting of the
  Association for Computational Linguistics (Volume 1: Long Papers), {ACL}
  2022, Dublin, Ireland, May 22-27, 2022}, {Smaranda Muresan}, {Preslav Nakov},
  {and} {Aline Villavicencio} (Eds.). Association for Computational
  Linguistics, 5180--5197.
\newblock
\showDOI{%
\url{http://dx.doi.org/10.18653/v1/2022.acl-long.356}}


\bibitem[\protect\citeauthoryear{Xu, Santu, and Veeramachaneni}{Xu
  et~al\mbox{.}}{2019}]%
        {xu2019mlfriend}
{Lei Xu}, {Shubhra Kanti~Karmaker Santu}, {and} {Kalyan Veeramachaneni}. 2019.
\newblock \showarticletitle{MLFriend: Interactive prediction task
  recommendation for event-driven time-series data}.
\newblock {\em arXiv preprint arXiv:1906.12348\/} (2019).
\newblock


\bibitem[\protect\citeauthoryear{Young, Xing, Pandelea, Ni, and Cambria}{Young
  et~al\mbox{.}}{2022}]%
        {young_fusing_2022}
{Tom Young}, {Frank Xing}, {Vlad Pandelea}, {Jinjie Ni}, {and} {Erik Cambria}.
  2022.
\newblock \showarticletitle{Fusing Task-Oriented and Open-Domain Dialogues in
  Conversational Agents}.
\newblock  {36}, 10 (2022), 11622--11629.
\newblock
\showISSN{2374-3468}
\showDOI{%
\url{http://dx.doi.org/10.1609/aaai.v36i10.21416}}
\newblock
\shownote{Number: 10.}


\bibitem[\protect\citeauthoryear{Zeng, Liu, Du, Wang, Lai, Ding, Yang, Xu,
  Zheng, Xia, Tam, Ma, Xue, Zhai, Chen, Zhang, Dong, and Tang}{Zeng
  et~al\mbox{.}}{2022}]%
        {GLM-130B}
{Aohan Zeng}, {Xiao Liu}, {Zhengxiao Du}, {Zihan Wang}, {Hanyu Lai}, {Ming
  Ding}, {Zhuoyi Yang}, {Yifan Xu}, {Wendi Zheng}, {Xiao Xia}, {Weng~Lam Tam},
  {Zixuan Ma}, {Yufei Xue}, {Jidong Zhai}, {Wenguang Chen}, {Peng Zhang},
  {Yuxiao Dong}, {and} {Jie Tang}. 2022.
\newblock \showarticletitle{{GLM-130B:} An Open Bilingual Pre-trained Model}.
\newblock {\em CoRR\/}  {abs/2210.02414} (2022).
\newblock
\showDOI{%
\url{http://dx.doi.org/10.48550/arXiv.2210.02414}}


\bibitem[\protect\citeauthoryear{Zhang, Lan, Pang, Guo, and Cheng}{Zhang
  et~al\mbox{.}}{2019}]%
        {ReCosa-context-detect}
{Hainan Zhang}, {Yanyan Lan}, {Liang Pang}, {Jiafeng Guo}, {and} {Xueqi Cheng}.
  2019.
\newblock \showarticletitle{ReCoSa: Detecting the Relevant Contexts with
  Self-Attention for Multi-turn Dialogue Generation}. In {\em Proceedings of
  the 57th Conference of the Association for Computational Linguistics, {ACL}
  2019, Florence, Italy, July 28- August 2, 2019, Volume 1: Long Papers}, {Anna
  Korhonen}, {David~R. Traum}, {and} {Llu{\'{\i}}s M{\`{a}}rquez} (Eds.).
  Association for Computational Linguistics, 3721--3730.
\newblock
\showDOI{%
\url{http://dx.doi.org/10.18653/v1/p19-1362}}


\bibitem[\protect\citeauthoryear{Zhang, Roller, Goyal, Artetxe, Chen, Chen,
  Dewan, Diab, Li, Lin, Mihaylov, Ott, Shleifer, Shuster, Simig, Koura,
  Sridhar, Wang, and Zettlemoyer}{Zhang et~al\mbox{.}}{2022}]%
        {opt}
{Susan Zhang}, {Stephen Roller}, {Naman Goyal}, {Mikel Artetxe}, {Moya Chen},
  {Shuohui Chen}, {Christopher Dewan}, {Mona~T. Diab}, {Xian Li}, {Xi~Victoria
  Lin}, {Todor Mihaylov}, {Myle Ott}, {Sam Shleifer}, {Kurt Shuster}, {Daniel
  Simig}, {Punit~Singh Koura}, {Anjali Sridhar}, {Tianlu Wang}, {and} {Luke
  Zettlemoyer}. 2022.
\newblock \showarticletitle{{OPT:} Open Pre-trained Transformer Language
  Models}.
\newblock {\em CoRR\/}  {abs/2205.01068} (2022).
\newblock
\showDOI{%
\url{http://dx.doi.org/10.48550/arXiv.2205.01068}}


\bibitem[\protect\citeauthoryear{Zhang, Sun, Galley, Chen, Brockett, Gao, Gao,
  Liu, and Dolan}{Zhang et~al\mbox{.}}{2020}]%
        {DialoGPT-sigdial}
{Yizhe Zhang}, {Siqi Sun}, {Michel Galley}, {Yen{-}Chun Chen}, {Chris
  Brockett}, {Xiang Gao}, {Jianfeng Gao}, {Jingjing Liu}, {and} {Bill Dolan}.
  2020.
\newblock \showarticletitle{{DIALOGPT} : Large-Scale Generative Pre-training
  for Conversational Response Generation}. In {\em Proceedings of the 58th
  Annual Meeting of the Association for Computational Linguistics: System
  Demonstrations, {ACL} 2020, Online, July 5-10, 2020}, {Asli Celikyilmaz}
  {and} {Tsung{-}Hsien Wen} (Eds.). Association for Computational Linguistics,
  270--278.
\newblock
\showDOI{%
\url{http://dx.doi.org/10.18653/v1/2020.acl-demos.30}}


\bibitem[\protect\citeauthoryear{Zoph and Le}{Zoph and Le}{2016}]%
        {zoph2016neural}
{Barret Zoph} {and} {Quoc~V Le}. 2016.
\newblock \showarticletitle{Neural architecture search with reinforcement
  learning}.
\newblock {\em arXiv preprint arXiv:1611.01578\/} (2016).
\newblock


\bibitem[\protect\citeauthoryear{Zoph, Vasudevan, Shlens, and Le}{Zoph
  et~al\mbox{.}}{2017}]%
        {zoph2017learning}
{Barret Zoph}, {Vijay Vasudevan}, {Jonathon Shlens}, {and} {Quoc~V Le}. 2017.
\newblock \showarticletitle{Learning transferable architectures for scalable
  image recognition}.
\newblock {\em arXiv preprint arXiv:1707.07012\/} {2}, 6 (2017).
\newblock


\end{thebibliography}
\bibliographystyle{ACM-Reference-Format-Journals}

\end{document}